\documentclass[10pt,twocolumn,letterpaper]{article}

\usepackage{cvpr}             
\usepackage{graphicx}
\usepackage{amsmath}
\usepackage{amssymb}
\usepackage{booktabs}

\usepackage{amsmath}

  \def\mD{{\mathcal D}}

  \def\mL{{\mathcal L}}

  \def\mO{{\mathcal O}}

  \def\mX{{\mathcal X}}

  \DeclareMathAlphabet\mathbfcal{OMS}{cmsy}{b}{n}

  \def\0{{\bf 0}}
  \def\1{{\bf 1}}

  \def\bx{{\bf x}}
  \def\by{{\bf y}}
  \def\bz{{\bf z}}

  \def\bx{{\bf x}}
  
  \def\by{{\bf y}}

  \def\bz{{\bf z}}

  \def\vtheta{{\bm{\theta}}}
  
  \DeclareMathOperator*{\argmax}{arg\,max}

\usepackage[ruled,vlined,linesnumbered]{algorithm2e}
\usepackage{threeparttable}
\usepackage{booktabs} 
\usepackage{multicol,multirow}
\usepackage{amsmath,amsfonts,bm}
\usepackage{xcolor}
\usepackage[switch]{lineno}
\usepackage{bbding}
\usepackage{balance}
\usepackage{xspace}

\newcommand*\samethanks[1][\value{footnote}]{\footnotemark[#1]}
\newcommand{\CLI}{CLI\xspace}

\def\ie{\emph{i.e.,}}
\def\eg{\emph{e.g.,}}

\def\ttt{\textcolor{black}}

\usepackage[pagebackref,breaklinks,colorlinks]{hyperref}

\usepackage[capitalize]{cleveref}
\crefname{section}{Sec.}{Secs.}
\Crefname{section}{Section}{Sections}
\Crefname{table}{Table}{Tables}
\crefname{table}{Tab.}{Tabs.}

\begin{document}

\title{Boost Test-Time Performance with Closed-Loop Inference}

\author{Shuaicheng Niu$^1$\thanks{First two authors contributed equally.} ~~ ~ Jiaxiang Wu$^2$\samethanks ~~ ~ Yifan Zhang$^3$ ~~ ~ Guanghui Xu$^1$ \\ Haokun Li$^1$ ~~ ~ Peilin Zhao$^2$ ~~ ~ Junzhou Huang$^2$ ~~ ~ Yaowei Wang$^4$ ~~ ~ Mingkui Tan$^1$\thanks{Corresponding author. E-mail: {\tt sensc@mail.scut.edu.cn}, {\tt  mingkuitan@scut.edu.cn}}
\\[0.152cm]
	$^1$South China University of Technology ~~ ~
	$^2$Tencent AI Lab \\
	$^3$National University of Singapore ~~ ~
	$^4$PengCheng Laboratory
}

\maketitle

\begin{abstract}
  Conventional deep models predict a test sample with a single forward propagation, which, however, may not be sufficient for predicting hard-classified samples. On the contrary, we human beings may need to carefully check the sample many times before making a final decision. During the recheck process, one may refine/adjust the prediction by referring to related samples. Motivated by this, we propose to predict those hard-classified test samples in a looped manner to boost the model performance. However, this idea may pose a critical challenge: how to construct looped inference, so that the original erroneous predictions on these hard test samples can be corrected with little additional effort. To address this, we propose a general Closed-Loop Inference (CLI) method. Specifically, we first devise a filtering criterion to identify those hard-classified test samples that need additional inference loops. For each hard sample, we construct an additional auxiliary learning task based on its original top-$K$ predictions to calibrate the model, and then use the calibrated model to obtain the final prediction. Promising results on ImageNet (in-distribution test samples) and ImageNet-C (out-of-distribution test samples) demonstrate the effectiveness of CLI in improving the performance of any pre-trained model.
\end{abstract}

\section{Introduction}

Deep neural networks have made promising progress in classification tasks~\cite{szegedy2017inceptionv4,dosovitskiy2021vit}, such as face recognition~\cite{wu2018coupled, Li_2021_CVPR} and video recognition~\cite{wang2018nonlocal,liu2020teinet}. In these tasks, existing methods often train a deep model through multiple epochs on a training set, and then predict test samples through a single forward propagation~\cite{he2016deep,huang2017densely}.  In this sense, the decision boundary of the pre-trained model remains constant regardless of the differences of test samples, which prevents the model from further improving its performance at test time.

In fact, the above method is quite different from the process of humans recognizing objects. For those images with simple objects, humans can quickly make predictions with high confidence. While for certain ambiguous objects, \eg~with occlusion or high similarity among different categories, humans often need to take a second thought to recognize. In the second thought, humans may compare the given object with some similar objects to determine the corresponding category. Inspired by this, given a pre-trained model, we seek to predict those hard-classified test samples in a looped manner to further boost the model performance. However, this raises the following key challenges.

\begin{figure}[t]
\centering
\includegraphics[width=0.8\linewidth]{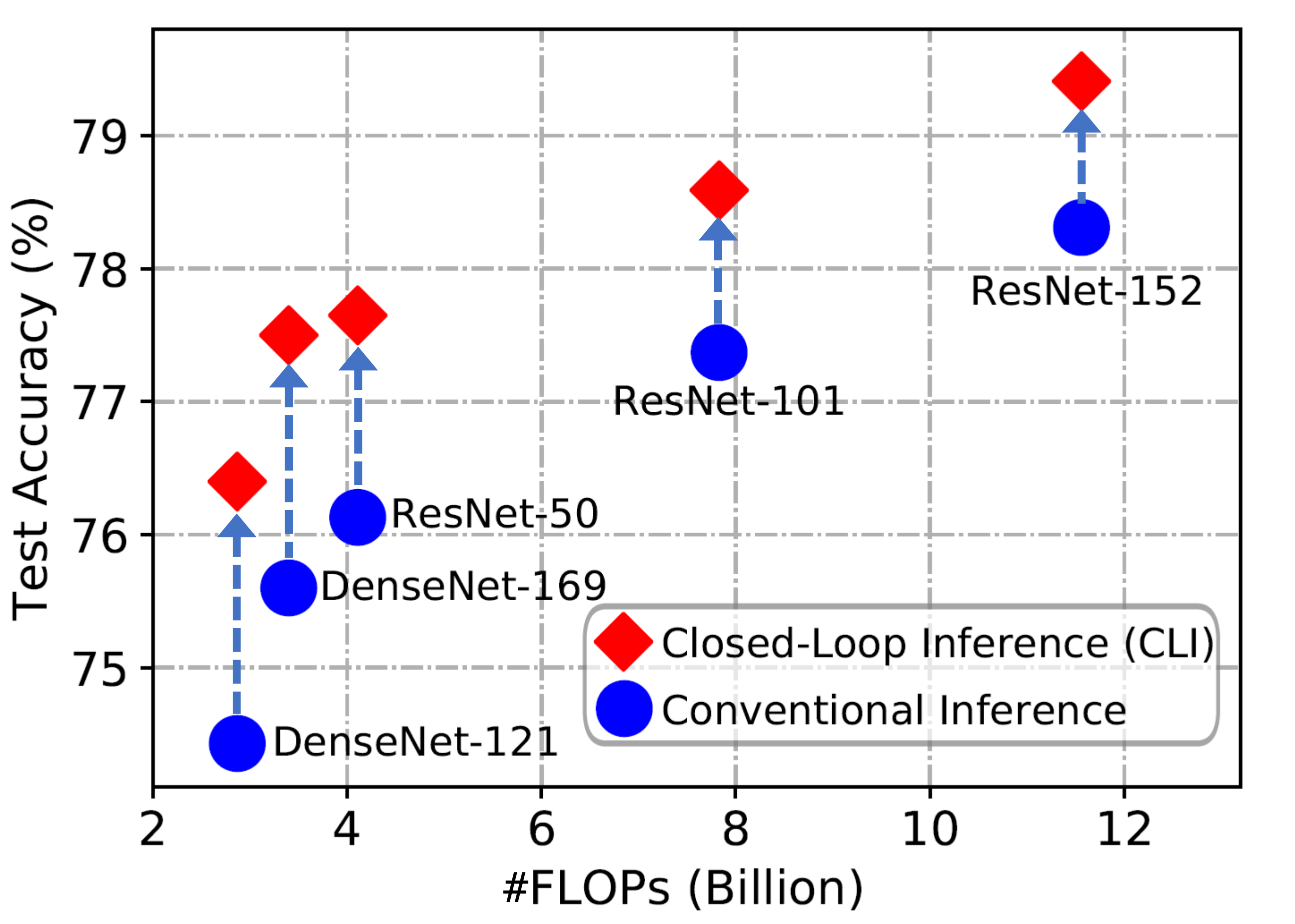}
\vspace{-0.1in}
\caption{Effectiveness of our Closed-Loop Inference (\CLI) over DenseNets and ResNets on ImageNet. 
\CLI helps to boost the accuracy of pre-trained models by conducting test-time learning for those hard-classified test samples.} 
\vspace{-0.05in}
\label{fig:illustration}
\end{figure}

\textbf{First}, distinguishing between easy-classified and hard-classified test samples is quite important, as performing a second-loop inference for easy test samples is unnecessary and increases the computational burden. However, in the context of our looped inference, this is still an open question. \textbf{Second}, at the second loop of predicting hard test samples, how to improve the model's predictive ability to enable it to correct those erroneous predictions is unclear. Test time learning-based methods~\cite{sun2020test,hansen2021self,mummadi2021test,chi2021test} exploit test samples to conduct self-supervised training (\eg~rotation prediction~\cite{gidaris2018unsupervised}) before making the final prediction. However, these methods are specially designed for overcoming the distribution shift between training and test data. When training and test data distributions are the same and the pre-trained model has been well trained on this distribution, self-supervised training may 1) fail to provide new information for model learning or 2) cause the well-trained model to deviate from its optimal solution, thereby having limited potentials to boost the model's predictive ability.

To address the above challenges, we propose a Closed-Loop Inference (\CLI) method, in which we conduct test-time auxiliary learning to boost the model's predictive ability for those hard-classified samples. To be specific, we exploit a score of maximum class probability \cite{hendrycks17baseline} to measure the prediction confidence, and then use this score to identify low-confident (hard) test samples. To perform auxiliary learning, we conducted preliminary studies and found that 1) the predicted top-$K$ classes for low-confident samples are often more difficult to classify and thus provide more fine-grained classification information (see the visualization of hard-classified images in Supplementary) and 2) the true category of low-confident samples often comes from the predicted top-$K$ classes. Therefore, we build an auxiliary training set by selecting samples of predicted top-$K$ classes from the entire training set, and then perform auxiliary training on it to further boost the model's classification ability on these fine-grained top-$K$ categories. Moreover, to improve the overall efficiency, we cluster low-confident samples via K-means clustering. After that, we perform auxiliary learning only once for each cluster and then predict all low-confident samples within this cluster.

In our experiments, we evaluate \CLI from two aspects. (1) We apply \CLI to various pre-trained networks (such as ResNet~\cite{he2016deep}, DenseNet~\cite{huang2017densely} and EfficientNet~\cite{tan2019efficientnet}) to boost the model performance on in-distribution test samples. Equipped with \CLI, the performance of pre-trained models are consistently improved, \eg~\CLI obtains 2.44\% and 1.90\% accuracy gain on ImageNet~\cite{deng2009imagenet} for ResNet-18 and DenseNet-169, respectively. (2) We incorporate \CLI into a previous SOTA method Test-Time Training~\cite{sun2020test} to further improve its generalization ability on out-of-distribution test samples. Empirical results with ResNet-50 on ImageNet-C~\cite{hendrycks2019benchmarking}   demonstrate the effectiveness of our approach.

In summary, our main contributions are as follows: 
\begin{itemize}
    \item We propose a general Closed-Loop Inference approach that not only boosts the performance of any pre-trained classification models on in-distribution test samples, but also can be applied to existing out-of-distribution generalization methods to further improve their generalization ability on out-of-distribution test samples.
    \item We exploit the maximum class probability to identify easy-classified test samples. For these samples, we make final predictions without additional inference loops, thereby avoiding unnecessary computation.
    \item For hard-classified test samples, we construct an auxiliary learning task at test time to boost the model's predictive ability on some local classification areas, so that the samples with erroneous original predictions can be corrected.
\end{itemize}

\begin{figure*}[t!]
\centering
\includegraphics[width=1.0\linewidth]{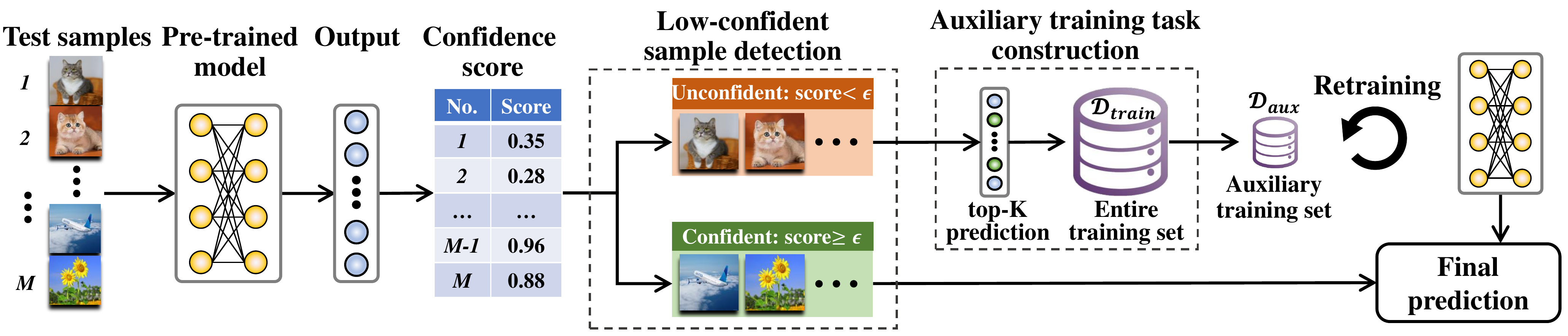}
\caption{An overall illustration of Closed-Loop Inference (\CLI). Given any pre-trained model and test samples, we first feed test samples to the model to obtain the corresponding outputs, and then compute the confidence score of each sample. For confident samples, we make the final prediction based on the current model's output. For unconfident samples, we will further conduct auxiliary training based on the model's top-$K$ predictions, and then use the updated model for the final prediction.
}
\label{fig:overall}
\end{figure*}

\section{Related Work}

\noindent\textbf{Test-time training}
(TTT)~\cite{sun2020test}
is the most recent work related to ours. Specifically, TTT trains a classification model using both a supervised learning objective and self-supervisions (rotation prediction~\cite{gidaris2018unsupervised}). Then, given a test sample, TTT will first train the model using this sample with self-supervisions and then use the updated model for final prediction. After that, the idea of TTT has been applied to many real-world applications, such as human pose estimation~\cite{hao2021test}, dynamic scene deblurring~\cite{chi2021test}, and long-tailed learning~\cite{zhang2021unleashing}. 
Compared with TTT that is designed for improving the model performance on out-of-distribution (OOD) test samples, our method is more general as it 1) not only can be applied to existing OOD generalization methods to further boost the performance 2) but also improves the predictive performance of any pre-trained model on test samples that have the same distribution with training samples. Moreover, TTT conducts auxiliary training via self-supervisions, while we seek to enhance the predictive performance by boosting the model's classification ability on some fine-grained classification areas (\ie~samples of top-$K$ classes) via supervised learning objectives.

~\\
\noindent\textbf{Unsupervised domain adaptation (UDA)}
seeks to learn a model on an unlabeled target domain by leveraging the knowledge from a well-labeled source domain, where the distribution shifts often exist between two domains~\cite{long2016unsupervised,long2018conditional,pei2018multi,saito2018maximum,wang2019transferable,yang2020bi}. 
To this end, one of the most popular techniques is to devise a domain discriminator to learn domain-invariant features in an adversarial manner, such as~\cite{pei2018multi} and ~\cite{saito2018maximum}. 
Unlike UDA that learns a universal model on all samples to alleviate the distribution shift, we seek to boost the model performance for those hard-classified samples by further conducting fine-grained model learning near some hard classification areas, and we do not assume that there must exist distribution shifts. Moreover, UDA tackles the whole target domain together since it requires joint training on the target domain. In this sense, when the target domain has very limited samples, UDA may get into fail, while \CLI still works even the target domain only has one sample.

~\\
\noindent\textbf{Dynamic inference (DI)} predicts different samples using data-dependent architectures or parameters, thereby improving the inference efficiency or the model's representation power~\cite{han2021dynamic}. 
Specifically, early existing methods allow samples (easy to classify) to be predicted using the early outputs of cascade DNNs~\cite{park2015big} or networks with multiple intermediate classifiers~\cite{guan2018energy}. Moreover, skipping methods selectively activate the model components, \eg~layers~\cite{guo2019dynamic}, branches~\cite{liu2018dynamic}, or sub-networks~\cite{cheng2020instanas} conditioned on the sample. Unlike DI to dynamically allocate computation for each sample, we seek to boost the performance of a static model on those samples with low-confident predictions.

\section{Closed-Loop Inference}

Let $f(\bx;\Theta)$ denote a model pre-trained on $\mD_{train}=\{(\bx_i,y_i)\}_{i=1}^{N}$, where $\bx_i\small{\in}\mX$, $y_i\small{\in}\{1,...,C\}$, $\mX$ is the input image space and $C$ is the number of classes. $\Theta\small{=}\{\vtheta_1,...\vtheta_L\}$ denotes the model's parameter set and $L$ denotes the number of layers.  Given test data $\mD_{test}=\{\bx_j\}_{j=1}^{M}$, existing methods often make predictions through a single forward propagation. For each $\bx\in\mD_{test}$, the predicted class is:
\begin{equation}\label{eq:conventional_inference}
    \hat{y} = \argmax_{y} f(y|\bx;\Theta), ~~\text{where}~~y\in\{1,...,C\}.
\end{equation}
In this way, the predictive performance of $f(\cdot;\Theta)$ will remain constant once the training is finished. In this work, we consider a challenging question: Can we break away from Eqn.~(\ref{eq:conventional_inference}) and boost the predictive ability of $f(\cdot;\Theta)$ at test time by sufficiently mining the knowledge contained in the model and test samples?

We start our answer from a simple observation. When humans recognize a hard-classified image, they may inspect the image repeatedly and compare the objects in this image with some other similar objects to determine the final prediction. Motivated by this, we propose a Closed-Loop Inference (\CLI) method that seeks to boost the model's predictive ability on some local classification area (\ie~a few categories that are most similar to the ground-truth) at test time for those hard-classified samples. In the context of \CLI, we address two key problems: 1) how to design a filtering criterion to identify hard-classified test samples that need additional loops; and 2) how to construct the looped inference to boost the model's predictive ability.

The overall pipeline of \CLI is shown in Figure~\ref{fig:overall}. We first devise a metric to compute the prediction confidences on $\mD_{test}$, and then split it into confident/easy-classified $\mD_{test}^{high}$ and low-confident/hard-classified $\mD_{test}^{low}$ via thresholding (c.f. Section~\ref{sec:selection}). Then, for $\bx\small{\in} \mD_{test}^{low}$, we construct an auxiliary training set $\mD_{aux}$ by choosing images of the original predicted top-$K$ classes from the entire training set, and then perform auxiliary training on $\mD_{aux}$. After the auxiliary training, we use the updated model to make the final prediction. Moreover, we further improve the overall efficiency of \CLI based on K-means clustering  (c.f. Section~\ref{sec:task_construct}). We summarize the details of \CLI in Algorithm~\ref{alg:overall}.

\subsection{Detection of Low-Confident Samples}\label{sec:selection}
To perform closed-loop inference, the first key step is to detect low-confident test samples that need auxiliary training to further calibrate the prediction model. Here, high-confident samples do not require auxiliary training, since if the model is very confident in its prediction regarding a given test sample $\bx$, we can directly take the current prediction as the final one. This also helps to avoid unnecessary loops and improve the algorithm efficiency. To detect these low-confident samples, we also expect that the detection will find more samples with wrong top-1 but correct top-$K$ predictions, which will enable our auxiliary training to correct the model regarding those wrong top-1 predictions.
Given all these, we define the predictive confidence score of a pre-trained model $f(\cdot;\Theta)$ as its maximum softmax outputs  regarding test data $\bx$:
\begin{equation}\label{eq:confidence}
    S(\bx;f) = \max_{c}\frac{\exp^{\hat{y}^c}}{\sum_{j=1}^{C}\exp^{\hat{y}^j}},
\end{equation}
where $\hat{y}^c$ is the $c$-th element of the model's output $\hat{\by}=f(\bx;\Theta)=[\hat{y}^1,\hat{y}^2,...,\hat{y}^C]$. The above confidence score is also called maximum class probability \cite{hendrycks17baseline} and one can also exploit other confidence scores, such as entropy and energy score~\cite{liu2020energy}.  Here, we choose the softmax-max score since it has a stronger ability to select samples whose top-1 is erroneous while top-$K$ is correct (see Figure~\ref{fig:ablations_sum} (d)). Based on the $S(\bx;f)$ and a given threshold $\epsilon$, we detect low-confident test samples by:
\begin{equation}\label{eq:ood_detector}
\mathbb{I}(\bx ; \epsilon, f)=\left\{\begin{array}{ll}
0 & \text { if } S(\bx; f) \geq \epsilon \\
1 & \text { if } S(\bx; f)<\epsilon
\end{array}.\right.
\end{equation}
From Eqn~(\ref{eq:ood_detector}), the samples in $\mD_{test}$ with $\mathbb{I}(\bx ; \epsilon, f)=0$ are called confident samples, denoted by $\mD_{test}^{high}$, for which we directly make final predictions using the original pre-trained model. For the samples with $\mathbb{I}(\bx ; \epsilon, f)=1$, which are low-confident (denoted by $\mD_{test}^{low}$), we need to update the model using an auxiliary training task (described as follows) and then make the final prediction with the updated model.

\begin{algorithm}[t]
    \SetAlgoLined
	\caption{The pipeline of closed-loop inference}
	\label{alg:overall}
	\LinesNumbered
    \KwIn{Pre-trained model $f(\cdot; \Theta_s,\Theta_d)$;   test samples $\mD_{test}\small{=}\{\bx_j\}_{j=1}^{M}$; training data $\mD_{train}$; $K;Q$.} 
    Let $\Theta_d^0=\Theta_d$. // \emph{Save initial parameters} \\
    Compute confidence for each $\bx\small{\in}\mD_{test}$ via Eqn.~(\ref{eq:confidence}). \\
    Split $\mD_{test}$ into $\mD_{test}^{low}$ and $\mD_{test}^{high}$using Eqn.~(\ref{eq:ood_detector}). \\
    Compute the top-1 prediction $\hat{y}$ for  $\bx\in\mD_{test}^{high}$.\\
    Cluster samples in $\mD_{test}^{low}$ to $Q$ clusters. // \emph{Eqn.~(\ref{eq:k_means_cluster})} \\
    \For{$q=1,2,...Q$}{
        Let $\Theta_d=\Theta_d^0$. // \emph{Rollback parameters}\\
        Compute the center $\bar{\by}$ for the $q$-th cluster $\mO_q$. \\
        Obtain the top-$K$ classes of $\bar{\by}$, namely top\small{-}$K(\bar{\by})$. \\
        Construct $\mD_{aux}$ based on top-$K$($\bar{\by}$). // \emph{Eqn.~(\ref{eq:d_aux})} \\
        Update $\Theta_d$ by optimizing Eqn.~(\ref{eq:finetune_objective}). \\
        Predict top-1 result $\hat{y}$ for all $\bx$ in the $q$-th cluster using the \textbf{updated} $f(\cdot;\Theta_s,\Theta_d)$. \\
    }
    \KwOut{The top-1 prediction results $\{\hat{y}_j\}_{j=1}^M$.}
\end{algorithm}

\subsection{Construction of Auxiliary Training Task}\label{sec:task_construct}
After detecting hard-classified test samples, the next question is how to use them to improve the model's predictive ability.  Our key insight is that the true category of a sample with a wrong top-1 prediction often comes from the predicted top-$K$ classes, considering that the top-$K$ accuracy (\eg~$K=5$) of deep models is often higher than the top-1 one. 
Inspired by this, we propose to explore the fine-grained top-$K$ classification information for boosting model performance, and develop a novel auxiliary learning task that trains the model to better classify the predicted top-$K$ classes of hard samples. Next, we will first introduce how  \CLI conducts the auxiliary learning, and then discuss how to accelerate the learning process.

\paragraph{Sample-level auxiliary learning.} Formally, given a low-confident test sample $\bx$ (detected by Eqn.~\ref{eq:ood_detector}) and its original model output $\hat{\by}$, we construct a dataset $\mD_{aux}$ for auxiliary training. To be specific, we select the samples of the original predicted top-$K$ classes (denoted by top-$K(\hat{\by})$) from the entire training set $\mD_{train}$:
\begin{equation}\label{eq:d_aux}
    \mD_{aux}=\{(\bx,y)|(\bx,y)\in\mD_{train} ~\text{and}~ y\in \text{top-}K(\hat{\by})\}.
\end{equation}
Here, the number of $K$ is a hyper-parameter. It is worth noting that the construction of $\mD_{aux}$ can also choose partial samples of the top-$K(\hat{\by})$ categories rather than all samples of them, which is a trade-off between the performance and efficiency (see Supplementary).
Based on $\mD_{aux}$, we then conduct an auxiliary classification task to boost the model's classification ability on the predicted top-$K$ classes. Details about the learning objective of the auxiliary task can be found in Section~\ref{sec:training_details}.

\paragraph{Cluster-level auxiliary learning.}\label{sec:batch_version}
In practice, we may have many low-confident samples to predict, where different samples may have the same or similar top-$K$ predictions, \ie~the auxiliary learning task for different samples is the same or similar. In this sense, performing auxiliary training for each sample separately is unnecessary. To improve the overall efficiency, we propose to cluster these low-confident samples into clusters, and construct the auxiliary training task for each cluster instead of each sample, \ie~auxiliary training once and predicting many. 

Formally, given a set of test samples $\mD_{test}\small{=}\{\bx_j\}_{j=1}^M$, we first detect samples with low confidence via the method in Section~\ref{sec:selection} and denote the low-confident samples by $\mD_{test}^{low}$. Then, we use the softmax values of model's outputs to group samples of $\mD_{test}^{low}$ to $Q$ clusters:   
\begin{equation}\label{eq:k_means_cluster}
    \{\mO_q\}_{q=1}^{Q}=\text{K-means}(\{\text{SoftMax}(f(\bx;\Theta))~|~\bx\small{\in}\mD_{test}^{low}\}).
\end{equation}
For each cluster, we use its center (the average of SoftMax outputs, denoted by $\bar{\by}$) to compute the top-$K$ predicted classes and then construct the auxiliary training set $\mD_{aux}$.

\subsection{Auxiliary Learning with Contrastive Loss}\label{sec:training_details}
Based on the auxiliary training set $\mD_{aux}$ built in Section~\ref{sec:task_construct}, we calibrate the model $f(\bx;\Theta)$ by optimizing both a cross-entropy loss $\mL_{ce}$ for classification and a supervised contrastive loss (SCL) $\mL_{scl}$~\cite{khosla2020supervised}. Here, the motivation of exploiting SCL is that the predicted top-$K$ classes regarding a top-1 erroneous sample are often fine-grained (see image visualizations in Supplementary). In this sense, the corresponding features of these classes' images may gather around the classification boundary and are hard to classify. To alleviate this, we use SCL to 
encourage the model to learn low-entropy features for a single class (\ie~high intra-class compactness) and high-entropy features for different classes (\ie~large inter-class separation degree)~\cite{zhang2021unleashing}.

\ttt{Given a training sample $\bx_i$, we feed it into the model to obtain the corresponding feature $\bz_i$ before the classifier and take $\bz_i$ as an anchor. Then, the contrastive loss takes the features from the same class to the anchor as positive pairs and those from the remaining classes as negative pairs. Assuming features are $\ell_2$-normalized, the supervised contrastive loss is computed by:}
\begin{equation}
    \mL_{scl}=-\frac{1}{n\left|P_{i}\right|} \sum_{i=1}^{n} \sum_{\bz_{j} \in P_{i}} \log \frac{\text{exp}^{\left(\bz_{i}^{\top} \bz_{j} / \tau\right)}}{\sum_{\bz_{k} \in A_{i}} \text{exp}^{\left(\bz_{i}^{\top} \bz_{k} / \tau\right)}}.
\end{equation}
\ttt{Here, $\tau$ is a temperature factor. $P_i$ and $A_i$ denote the positive pair set and the full pair set of the anchor $\bz_i$.} 
Then, the optimization formulation for auxiliary training is:
\begin{equation}\label{eq:finetune_objective}
\min_{\Theta_d} \mathbb{E}_{(\bx,y)\sim \mD_{aux}}[\mL_{ce}(\bx,y;\Theta_s, \Theta_d)+\lambda \mL_{scl}(\bx,y;\Theta_s, \Theta_d)],
\end{equation}
where $\Theta_s=\{\vtheta_1,...,\vtheta_H\}$, $\Theta_d=\{\vtheta_{H+1},...,\vtheta_L\}\subset\Theta$ denote the parameters of \textbf{s}hadow layers (extract general features) and \textbf{d}eep layers (extract specific features), respectively. In the auxiliary training process, we can only optimize the deep layers $\Theta_{d}$ and freeze $\Theta_{s}$. The reason is that training $\Theta_d$ is enough to achieve the same or higher performance than training the entire network (see Section~\ref{sec:ablations_all}). Moreover, this also helps to accelerate the auxiliary training process and reduce GPU memory consumption.

\section{Experiments}

We evaluate our Closed-Loop Inference (\CLI) method from two aspects: improving the test-time performance on (i) in-distribution test samples and (ii) out-of-distribution test samples. The source code will be released upon acceptance.

\paragraph{Implementation details.} 
\CLI is implemented as follows. The threshold $\epsilon$ in Eqn.~(\ref{eq:ood_detector}) for low-confident sample detection is set to 0.7 and 0.6 for in- and out-of- distribution experiments, respectively. We simply set the cluster number $Q$ of   K-means clustering to 400 on  ImageNet, the trade-off parameter $\lambda$ in Eqn.~(\ref{eq:finetune_objective}) to 1 and the temperature $\tau$ in Eqn.~(\ref{eq:finetune_objective}) to 0.07. The class number $K$ for auxiliary training set construction in Eqn.~(\ref{eq:d_aux}) is set to 10. For each auxiliary training, we train the entire model for 5 epochs with a batch size of 256. We apply an SGD optimizer with a weight decay of $10^{-4}$ and a momentum of 0.9. We initialize the learning rate as $0.01$ and decrease it by cosine annealing~\cite{SGDR}. We use random crop and horizontal flip for data augmentation.
All pre-trained models used in our experiments are obtained from the official GitHub of PyTorch\footnote{https://github.com/pytorch/vision/tree/master/torchvision/models.} or EfficientNet\footnote{https://github.com/lukemelas/EfficientNet-PyTorch.}.

\subsection{Results on In-Distribution Test Samples}
\paragraph{Dataset and networks.} 
We conduct all experiments on the ImageNet dataset~\cite{deng2009imagenet}. We apply \CLI to the following networks pre-trained on ImageNet: 1) human-designed heavy networks: ResNets~\cite{he2016deep}, ResNeXt~\cite{xie2017aggregated}, DenseNets~\cite{huang2017densely}, and Inception-V3~\cite{szegedy2016rethinking}; 2) human-designed compact networks: ShuffleNet~\cite{ma2018shufflenet}, MobileNetV2~\cite{sandler2018mobilenetv2}, and MobileNetV3~\cite{howard2019searching}; 3) automatically searched compact networks: EfficientNets~\cite{tan2019efficientnet}. Moreover, we also incorporate our \CLI into ResNeXt-WSL~\cite{mahajan2018exploring}, which is pre-trained on 940 million public images in a weakly-supervised fashion, followed by fine-tuning on ImageNet.

\begin{table}[t!]
\newcommand{\tabincell}[2]{\begin{tabular}{@{}#1@{}}#2\end{tabular}}
 \begin{center}
 \begin{threeparttable}
    \resizebox{1.0\linewidth}{!}{
 	\begin{tabular}{l|cc|c|c}\toprule
 	Model  & \#Params. & \#FLOPs & Original  & \CLI  (ours) \\
        \midrule
         ResNet-18~\cite{he2016deep} & 11.7M & 1.82B & 69.76 & \textbf{72.20 (+2.44)}\\
         ResNet-50~\cite{he2016deep} & 25.6M & 4.11B & 76.13 & \textbf{77.65 (+1.52)}\\
         ResNet-101~\cite{he2016deep} & 44.6M & 7.83B & 77.37 & \textbf{78.59 (+1.22)}\\
         ResNet-152~\cite{he2016deep} & 60.2M & 11.56B & 78.31 & \textbf{79.41 (+1.10)}\\
         ResNeXt-101~\cite{xie2017aggregated} & 88.8M & 16.48B & 79.31 & \textbf{79.89 (+0.58)}\\
         R-101-WSL~\cite{mahajan2018exploring} & 88.8M & 16.48B & 82.69 & \textbf{83.27 (+0.58)}\\
         \midrule
         DenseNet-121~\cite{huang2017densely} & 8.0M & 2.87B & 74.43 & \textbf{76.40 (+1.96)}\\
         DenseNet-169~\cite{huang2017densely} & 14.2M & 3.40B & 75.60 & \textbf{77.50 (+1.90)}\\
         Inception-V3~\cite{szegedy2016rethinking} & 23.8M & 5.73B & 77.47 & \textbf{78.84 (+1.37)}\\ 
         \midrule
         ShuffleNet-V2~\cite{ma2018shufflenet} & 2.3M & 0.15B & 69.36 & \textbf{70.37 (+1.01)}\\ 
        MobileNet-V2~\cite{sandler2018mobilenetv2} & 3.5M & 0.31B & 71.88 & \textbf{73.73 (+1.85)}\\ 
         MobileNet-V3~\cite{howard2019searching} & 5.5M & 0.23B & 74.04 & \textbf{75.21 (+1.17)}\\ 
         EfficientNet-B0~\cite{tan2019efficientnet} & 5.3M & 0.39B & 76.13 & \textbf{77.91 (+1.48)}\\
         EfficientNet-B1~\cite{tan2019efficientnet} & 7.8M & 0.70B & 78.38 & \textbf{79.28 (+0.90)}\\
        \bottomrule 
	\end{tabular}
	}
	 \end{threeparttable}
	 \end{center}
	 \vspace{-0.15in}
    \caption{Test accuracy (\%) on several classic and state-of-the-art models on ImageNet. R-101-WSL is short for ResNeXt-101-WSL. A detailed version of this Table that contains the number of images whose top-1 predictions are adjusted (by \CLI) from  False-to-True and True-to-False can be found in Supplementary.
    }
    \vspace{-0.1in}
	 \label{tab:main_results}
\end{table}

\paragraph{\CLI on various networks.} 
As shown in Table~\ref{tab:main_results}, our \CLI consistently outperforms Conventional Inference on various neural architectures.
According to the results, we have the following main observations.
\textbf{(1)} From ResNet-18 to ResNet-152, the performance gain obtained by \CLI decreases.
The reason is that the potential of \CLI to boost performance will decrease as the original performance of the considered model increases.
In other words, if a given predictive model has 100\% accuracy, the auxiliary training will be useless.
Nevertheless, when applying \CLI to a heavy and high-performance model, \ie~ResNeXt-101-WSL (82.69\% top-1 accuracy), we still achieve 0.58\% accuracy improvement. 
\textbf{(2)} The performance gain of \CLI not only depends on the model's original accuracy, but also relies on the neural architecture.
For example, although the original accuracy of DenseNet-169 is better than ShuffleNet-V2 (75.60\% v.s.~69.36\%),   the accuracy improvement of \CLI on DenseNet-169 is still larger than ShuffleNet-V2 (1.90\% v.s. 1.01\%).
\textbf{(3)} Equipped with \CLI, the performance of a compact model is able to surpass a heavy model.
For example, the \CLI accuracy of ResNet-50 is 77.65\%, which outperforms the original accuracy of ResNet-101 (77.37\%).

\paragraph{Comparisons with test-time training (TTT)~\cite{sun2020test}.}
We further compare \CLI with TTT based on ResNet-18 and ResNet-50 under different training epochs.
As shown in Table~\ref{tab:comparison_ttt}, TTT improves the predictive performance at the early training stage (\ie~20 and 40 epochs), while hampers the performance when the model goes to converge (\ie~60 and 90 epochs). The reason is that TTT is designed to overcome the distribution shift, rather than improve the predictive performance under the same domain. More specifically, when training and testing domains are from the same distribution which has been well learned by the pre-trained model, the test-time learning of TTT cannot provide new information to benefit the model learning. Compared with TTT, \CLI continuously improves the performance under all the training epochs, demonstrating its effectiveness.   

\begin{table}[t]
\newcommand{\tabincell}[2]{\begin{tabular}{@{}#1@{}}#2\end{tabular}}
 \begin{center}
 \begin{threeparttable}
    \resizebox{1.0\linewidth}{!}{
 	\begin{tabular}{c|c|c|cc}\toprule
 	Model & Epochs & Original Acc. & TTT & \CLI (ours) \\
 	\midrule
 	\multirow{4}{*}{\tabincell{c}{ResNet-18}}
 	     & 20 & 46.70\% & 48.86\%  & \textbf{53.45\%}  \\
 	     & 40 & 62.37\% & 62.80\%  & \textbf{66.73\%}  \\
 	     & 60 & 66.74\% & 66.38\%  & \textbf{69.92\%}  \\
 	     & 90 & 68.78\% & 67.70\%  & \textbf{71.33\%}  \\
    \midrule
 	\multirow{4}{*}{\tabincell{c}{ResNet-50}}
 	     & 20 & 47.17\% & 48.36\%  & \textbf{53.97\%}  \\
         & 40 & 63.86\% & 64.44\%  & \textbf{68.85\%}  \\
         & 60 & 70.45\% & 70.13\%  & \textbf{72.79\%} \\
         & 90 & 74.40\% & 73.25\%  & \textbf{75.96\%} \\
        \bottomrule
	\end{tabular}
	}
	 \end{threeparttable}
	 \end{center}
	 \vspace{-0.15in}
    \caption{Comparison with Test-Time Training (TTT)~\cite{sun2020test} on ResNet models trained on ImageNet. 
    The original models are trained using TTT's official source code.
    }
    \label{tab:comparison_ttt}
\end{table}

\begin{table}[t]
\newcommand{\tabincell}[2]{\begin{tabular}{@{}#1@{}}#2\end{tabular}}
 \begin{center}
 \begin{threeparttable}
    \resizebox{1.0\linewidth}{!}{
 	\begin{tabular}{c|cc}\toprule
 	Method & ResNet-18 (69.76\%) & ResNet-50 (76.13\%) \\
 	\midrule
          Pure Fine-Tuning & 70.14\% (+0.38\%) & 76.61\% (+0.48\%)\\
          \CLI (ours) & \textbf{72.20\% (+2.44\%)} & \textbf{77.65\% (+1.52\%)} \\
    \bottomrule
	\end{tabular}
	}
	 \end{threeparttable}
	 \end{center}
	 \vspace{-0.15in}
\caption{Comparison with Pure Fine-Tuning on ImageNet.
}
\label{tab:comparison_joint_training}
\end{table}

\paragraph{Comparison with pure fine-tuning (PFT).} To verify that the accuracy gain obtained by \CLI is not only from the additional training, we compare it with a baseline of PFT, which fine-tunes the pre-trained model on the joint of all categories' samples (entire training set). For fair comparisons, we train the model for 20 epochs to make the update iterations of PFT the same as \CLI. From Table~\ref{tab:comparison_joint_training}, \CLI achieves larger performance gains than PFT, \eg~2.44\% v.s. 0.38\% on ResNet-18. These results further verify the effectiveness of our auxiliary training strategy that exploits the top-$K$ prediction to further boost the top-1 performance.

\begin{table*}[t]
\newcommand{\tabincell}[2]{\begin{tabular}{@{}#1@{}}#2\end{tabular}}
 \begin{center}
 \begin{threeparttable}
    \resizebox{1.0\linewidth}{!}{
 	\begin{tabular}{c|l|ccc|cccc|cccc|cccc}
 	\multicolumn{2}{c}{} & \multicolumn{3}{c}{Noise} & \multicolumn{4}{c}{Blur} & \multicolumn{4}{c}{Weather} & \multicolumn{4}{c}{Digital} \\
 	\toprule
 	Severity & Method & Gauss. & Shot & Impulse & Defocus & Glass & Motion & Zoom & Snow & Frost & Fog & Britght & Contrast & Elastic & Pixel & JPEG \\
 	\midrule
 	\multirow{7}{*}{\tabincell{c}{Level-1}}
        & ResNet-50~\cite{he2016deep}  (base)       & 59.89&	57.87&	48.08&	55.89&	52.68&	62.01&	51.80&	52.13&	57.89&	58.04&	71.06&	61.23&	65.65&	63.51&	63.62 \\
        & AugMix~\cite{hendrycks2019augmix}           & 66.35& 65.58&	60.06&	\textbf{63.12}&	59.70&	\textbf{69.67}&	\textbf{60.93}&	\underline{\textbf{59.03}} &	62.48&	61.84&	\underline{\textbf{73.60}}&	68.75&	69.29&	68.85&	66.18 \\
        & Fast-AutoAugment~\cite{lim2019fast} & 64.82&	63.70&	54.66&	59.64&	61.67&	\underline{\textbf{65.48}} &	53.01&	55.91&	61.60&	65.44&	73.38&	69.28&	68.22&	66.27&	65.22 \\
        & ANT 3$\times3$~\cite{rusak2020simple}   & \textbf{68.51}&	\textbf{67.91}&	\textbf{65.13}&	60.07&	59.96&	\underline{\textbf{65.48}}&	55.07&	57.21&	62.36&	59.16&	72.50&	63.95&	67.15&	67.99&	\underline{\textbf{67.27}} \\
        & SIN~\cite{geirhos2018imagenet}              & 49.95&	46.58&	43.04&	37.63&	47.21&	49.36&	37.18&	47.10&	49.59&	53.33&	58.83&	55.62&	52.80&	56.68&	53.51 \\
        & TTT~\cite{sun2020test}              & 66.81&	66.83&	63.43&	61.77&	\underline{\textbf{63.71}} &	63.23&	58.42&	57.43&	\underline{\textbf{62.87}} &	\underline{\textbf{68.28}} &	73.21&	\underline{\textbf{69.98}} &	\underline{\textbf{69.31}} &	\underline{\textbf{69.70}} &	66.54 \\
        & TTT+\CLI (ours)  & \underline{\textbf{68.25}}   &	\underline{\textbf{67.79}}&	\underline{\textbf{63.99}} &	\underline{\textbf{62.36}} &	\textbf{64.96}&	65.23&	\underline{\textbf{60.05}}&	\textbf{59.10}&	\textbf{64.84}&	\textbf{69.67}&	\textbf{74.43}&	\textbf{71.26}&	\textbf{70.45}&	\textbf{70.91}&	\textbf{68.43} \\
    \midrule
 	\multirow{7}{*}{\tabincell{c}{Level-2}}
        & ResNet-50~\cite{he2016deep} (base)       & 48.21&	43.91&	39.01&	48.07&	38.69&	50.51&	42.16&	31.02&	40.40&	51.56&	69.30&	54.19&	43.88&	61.01&	60.37 \\
        & AugMix~\cite{hendrycks2019augmix}           & 58.63&	56.82&	51.64&	\textbf{58.49}&	47.77&	\textbf{64.89}&	53.77&	39.55&	47.16&	55.90&	71.94&	65.16&	47.40&	66.94&	63.31 \\
        & Fast-AutoAugment~\cite{lim2019fast} & 56.95&	54.94&	48.39&	52.70&	\textbf{49.54}&	55.45&	42.46&	37.19&	45.39&	60.92&	\underline{\textbf{72.39}}&	66.25&	45.74&	63.58&	61.96 \\
        & ANT 3$\times3$~\cite{rusak2020simple}   & \textbf{67.04}&	\textbf{65.47}&	\textbf{63.64}&	53.83&	\underline{\textbf{48.97}}&	\underline{\textbf{55.98}} &	45.51&	39.21&	48.01&	52.37&	70.64&	57.60&	45.49&	66.25&	\textbf{65.11} \\
        & SIN~\cite{geirhos2018imagenet}              & 43.46&	39.10&	36.94&	30.48&	37.20&	42.47&	30.06&	36.69&	39.97&	50.89&	57.32&	53.36&	35.06&	56.05&	49.32 \\
        & TTT~\cite{sun2020test}              & 62.26&	61.60&	58.37&	55.30&	47.76&	54.23&	\textbf{55.63}&	\underline{\textbf{48.83}} &	\underline{\textbf{48.53}} &	\underline{\textbf{66.10}} &	72.33&	\underline{\textbf{68.32}} &	\underline{\textbf{55.56}} &	\underline{\textbf{68.69}} &	63.46 \\
        & TTT+\CLI (ours)  & \underline{\textbf{62.93}} &	\underline{\textbf{62.17}} &	\underline{\textbf{58.67}}&	\underline{\textbf{55.51}}&	48.88&	55.82&	\underline{\textbf{55.51}}&	\textbf{49.00}&	\textbf{49.10}&	\textbf{67.38}&	\textbf{73.46}&	\textbf{69.39}&	\textbf{55.67}&	\textbf{69.84}&	\underline{\textbf{65.03}} \\
    \midrule
 	\multirow{7}{*}{\tabincell{c}{Level-3}}
 	     & ResNet-50~\cite{he2016deep} (base)      & 31.00&	28.54&	31.03&	33.03&	16.03&	33.32&	34.49&	33.57&	29.17&	42.65&	66.40&	40.94&	52.41&	49.09&	57.39 \\
         & AugMix~\cite{hendrycks2019augmix}           & 45.53&	45.18&	44.46&	\textbf{46.54}&	24.85&	\textbf{53.78}&	48.98&	41.01&	\underline{\textbf{35.92}} &	46.75&	69.13&	58.10&	60.23&	59.75&	61.09 \\
         & Fast-AutoAugment~\cite{lim2019fast} & 44.80&	43.79&	43.14&	37.75&	22.03&	38.63&	34.97&	39.43&	34.29&	53.01&	70.77&	59.32&	56.05&	53.55&	59.50 \\
         & ANT 3$\times3$~\cite{rusak2020simple}   & \textbf{62.75}&	\textbf{61.51}&	\textbf{61.40}&	40.89&	25.61&	\underline{\textbf{39.31}} &	37.57&	36.43&	\textbf{37.41}&	42.69&	67.84&	45.13&	57.69&	57.64&	\textbf{63.22} \\
         & SIN~\cite{geirhos2018imagenet}              & 36.01&	32.86&	33.03&	23.42&	23.07&	33.98&	28.52&	37.99&	33.45&	46.55&	55.35&	49.34&	54.67&	48.54&	46.16 \\
         & TTT~\cite{sun2020test}              & \underline{\textbf{55.00}} &	\underline{\textbf{55.19}} &	\underline{\textbf{54.30}} &	\underline{\textbf{46.11}} &	\textbf{32.54} &	36.79&	\textbf{52.60}&	\underline{\textbf{49.26}} &	34.97&	\underline{\textbf{63.63}} &	\underline{\textbf{70.78}} &	\underline{\textbf{64.93}} &	\underline{\textbf{66.69}} &	\underline{\textbf{63.98}} &	61.43 \\
         & TTT+\CLI (ours)  & 54.85&	54.21&	53.76&	44.76&	\underline{\textbf{31.07}} &	37.33&	\underline{\textbf{52.43}} &	\textbf{49.54}&	35.02&	\textbf{64.55}&	\textbf{71.97}&	\textbf{65.91}&	\textbf{67.62}&	\textbf{65.02}&	\underline{\textbf{63.06}} \\
        \bottomrule
	\end{tabular}
	}
	 \end{threeparttable}
	 \end{center}
	 \vspace{-0.15in}
    \caption{Test accuracy (\%) on ImageNet-C with different severity-levels. The backbone model for all compared methods and ours is ResNet-50. The \textbf{bold} number indicates the best result and \underline{\textbf{bold}} number with an underline indicates the second best result.
    }
    \label{tab:comparison_oodg}
\end{table*}

\subsection{Results on Out-of-Distribution Test Samples}
\paragraph{Dataset.} We also evaluate \CLI by testing whether it improves the out-of-distribution (OOD) performance of models on ImageNet-C~\cite{hendrycks2019benchmarking}, which is constructed by corrupting the  ImageNet~\cite{deng2009imagenet} test set. The corruption consists of 15 noise, blur, weather, and digital types, each appearing at 5 different severity levels. More details are put in Appendix.

\paragraph{Compared methods.}  We compare \CLI with following techniques that improve OOD performance. AugMix~\cite{hendrycks2019augmix} mixes multiple augmented images and then exploits a Jensen-Shannon Divergence consistency loss for training. Fast-AutoAugment~\cite{lim2019fast} is an auto-searched combination of data augmentation policies. ANT~\cite{rusak2020simple} introduces Gaussian, Speckle, and Adversarial noise into the model training. SIN~\cite{geirhos2018imagenet} jointly trains a model on both original ImageNet images and Stylized-ImageNet images.  Test-Time Training (TTT)~\cite{sun2020test} performs additional self-supervised learning at test time, different from the above methods that anticipate the distribution shifts. In this section, we apply \CLI to TTT to further improve its performance (see Supplementary for more implementation details). We use the pre-trained model obtained from the baselines'  GitHub for testing.

\paragraph{Comparison.} From Table~\ref{tab:comparison_oodg}, we have the following main observations. First, equipped with our \CLI, the performances of TTT are further improved, achieving the best or comparable performance on most corruption types, which verifies the effectiveness of \CLI on OOD tasks. Second, ANT achieves the best performance on the Noisy corruption, since it introduces the Gaussian and Adversarial noise into training, which anticipates the distribution shift between training and test domains. Nevertheless, when the severity of the noisy level is slight (\ie~level=1), TTT+\CLI still achieves comparable performance compared with ANT. Third, when the corruption level becomes more severe (\eg~level-3 of Gaussian noise), \CLI fails to further boost TTT. The reason is that the distribution of our auxiliary training set is too far from the test distribution. In this sense, the auxiliary training may tend to overfit the clean data and cannot generalize well to corrupted test data. To conquer this, one possible solution is combining \CLI with ANT, in which noisy augmentations will be directly introduced into the training phase and thus compensate for the severe distribution shift.

\subsection{Ablations}\label{sec:ablations_all}

In this section, we conduct ablations to verify the effects of each hyper-parameter and component in \CLI. 
For each experiment, we only adjust one hyper-parameter or component and keep others fixed.
All experiments are conducted on ImageNet pre-trained ResNet-18 or ResNet-50.

\paragraph{Metrics.} 1) \textbf{\#F2T}: the number of test images whose top-1 prediction is corrected from \textbf{f}alse \textbf{t}o \textbf{t}rue (F2T). 2) \textbf{\#T2F}: the number of test images whose top-1 is corrected from \textbf{t}rue \textbf{t}o \textbf{f}alse (T2F). The percentage (\#\%) for F2T/T2F is calculated by F2T/T2F dividing the total number of test samples (50,000 for ImageNet). 
3) \textbf{Time per Image (TPI)}: the average run-time for testing one image via \CLI, calculated by the total run time (on a single Tesla-V100) of our algorithm dividing the number of test samples (50,000).

\paragraph{Supervised contrastive loss (SCL)~\cite{khosla2020supervised}.}
As shown in Table~\ref{tab:ablation_scl}, equipped with SCL, \CLI further improves \#F2T images while keeping \#T2F images comparable, boosting the overall performance. These results verify our motivation in Section~\ref{sec:training_details} that SCL helps to learn features with high intra-class compactness and large inter-class separation degree, thereby improving the performance on fine-grained classification areas (\ie~hard-classified samples that are near the decision boundary). It is worth noting that although \CLI is able to correct the prediction of test images from false to true, it will also adjust a smaller number of predictions from true to false. One possible reason is that the auxiliary training makes the model forget some previous knowledge. How to reduce \#T2F is still an open question, and we leave it to our future work.

\begin{table}[t]
\newcommand{\tabincell}[2]{\begin{tabular}{@{}#1@{}}#2\end{tabular}}
 \begin{center}
 \begin{threeparttable}
    \resizebox{1.0\linewidth}{!}{
 	\begin{tabular}{c|c|cc|c}\toprule
 	Model & SCL &  \#F2T ($\uparrow$) & \#T2F ($\downarrow$) & \CLI Acc. \\
 	\midrule
 	\multirow{2}{*}{\tabincell{c}{ResNet-18 \\ (69.76\%)}}
 	     & $\times$ & 2,271 (4.54\%) & 1,269 (2.54\%) & 71.76\% (+2.00\%) \\
 	     & $\checkmark$ & 2,491 (4.98\%) & 1,269 (2.54\%) & \textbf{72.20\% (+2.44\%)}  \\
    \midrule
 	\multirow{2}{*}{\tabincell{c}{ResNet-50 \\ (76.13\%)}}
 	     & $\times$ & 1,799 (3.60\%) & 1,128 (2.26\%)  & 77.47\% (+1.34\%)  \\
         & $\checkmark$ & 1,925 (3.85\%) & 1,165 (2.33\%) & \textbf{77.65\% (+1.52\%)} \\
        \bottomrule
	\end{tabular}
	}
	 \end{threeparttable}
	 \end{center}
\vspace{-0.15in}
\caption{Ablation on supervised contrastive loss (SCL) in \CLI. Experiments are conducted on ImageNet.
}
\label{tab:ablation_scl}
\end{table}

\begin{figure*}[!h]
  \centering
  \begin{subfigure}{0.25\linewidth} 
    \includegraphics[width=1.0\linewidth]{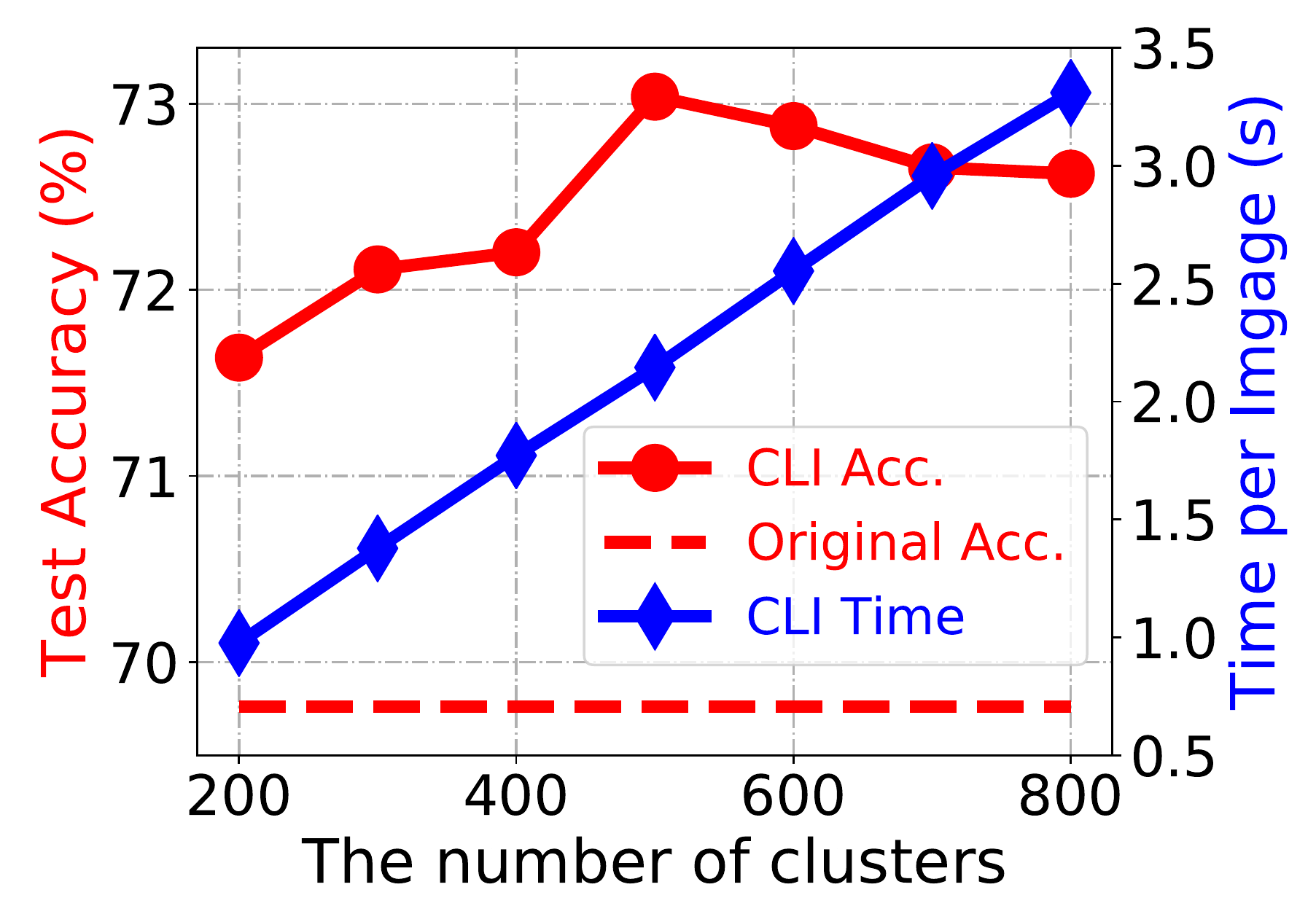}
    \caption{Effects of cluster number $Q$.}
    \label{fig:ablation_num_clusters}
  \end{subfigure}
  \hspace{-0.1in}
  \begin{subfigure}{0.25\linewidth}
    \includegraphics[width=1.0\linewidth]{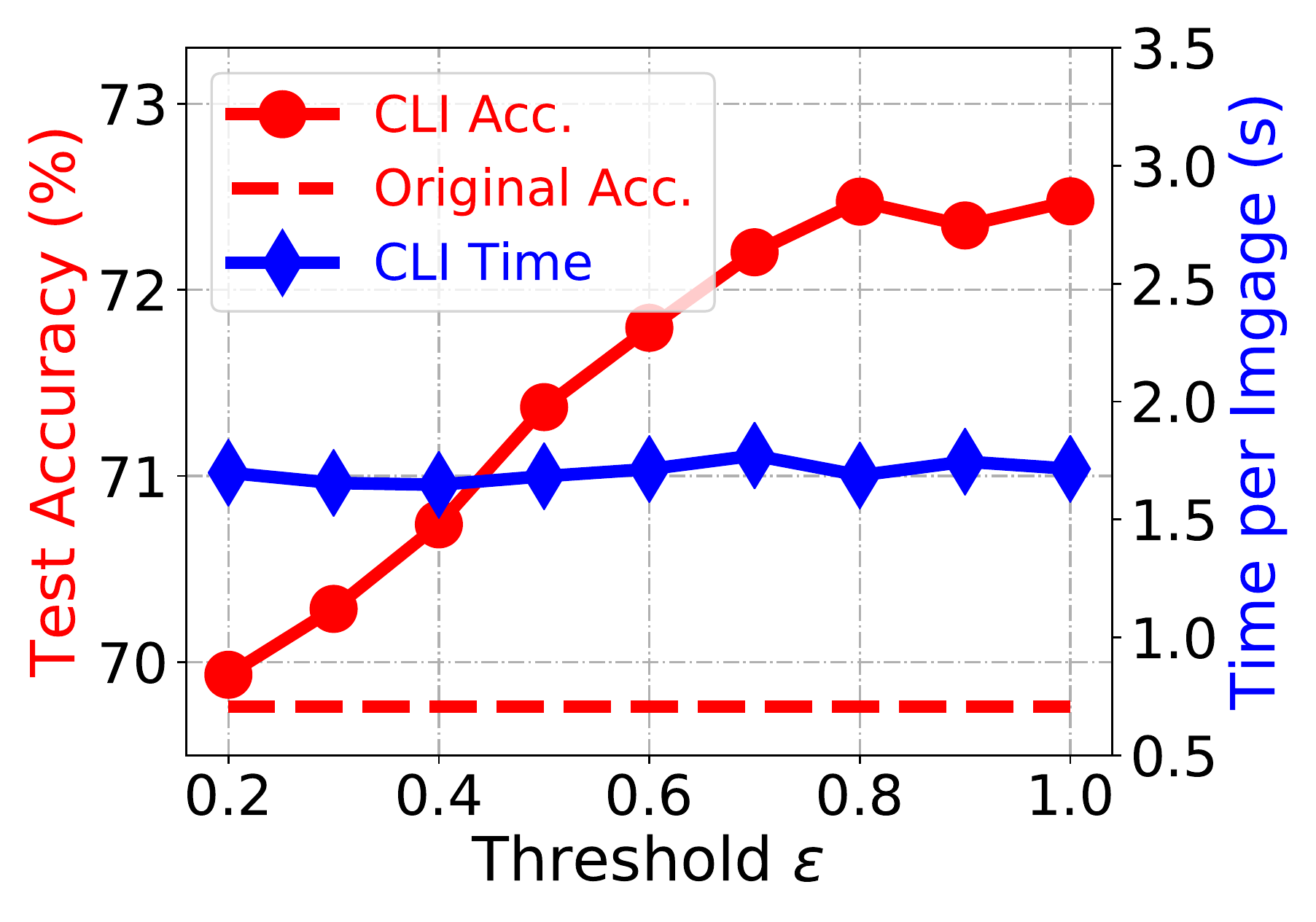}
    \caption{Effects of thresholds $\epsilon$ in Eqn.~(\ref{eq:ood_detector}).}
    \label{fig:ablation_thresholds}
  \end{subfigure}
  \hspace{-0.1in}
  \begin{subfigure}{0.25\linewidth}
    \includegraphics[width=1.0\linewidth]{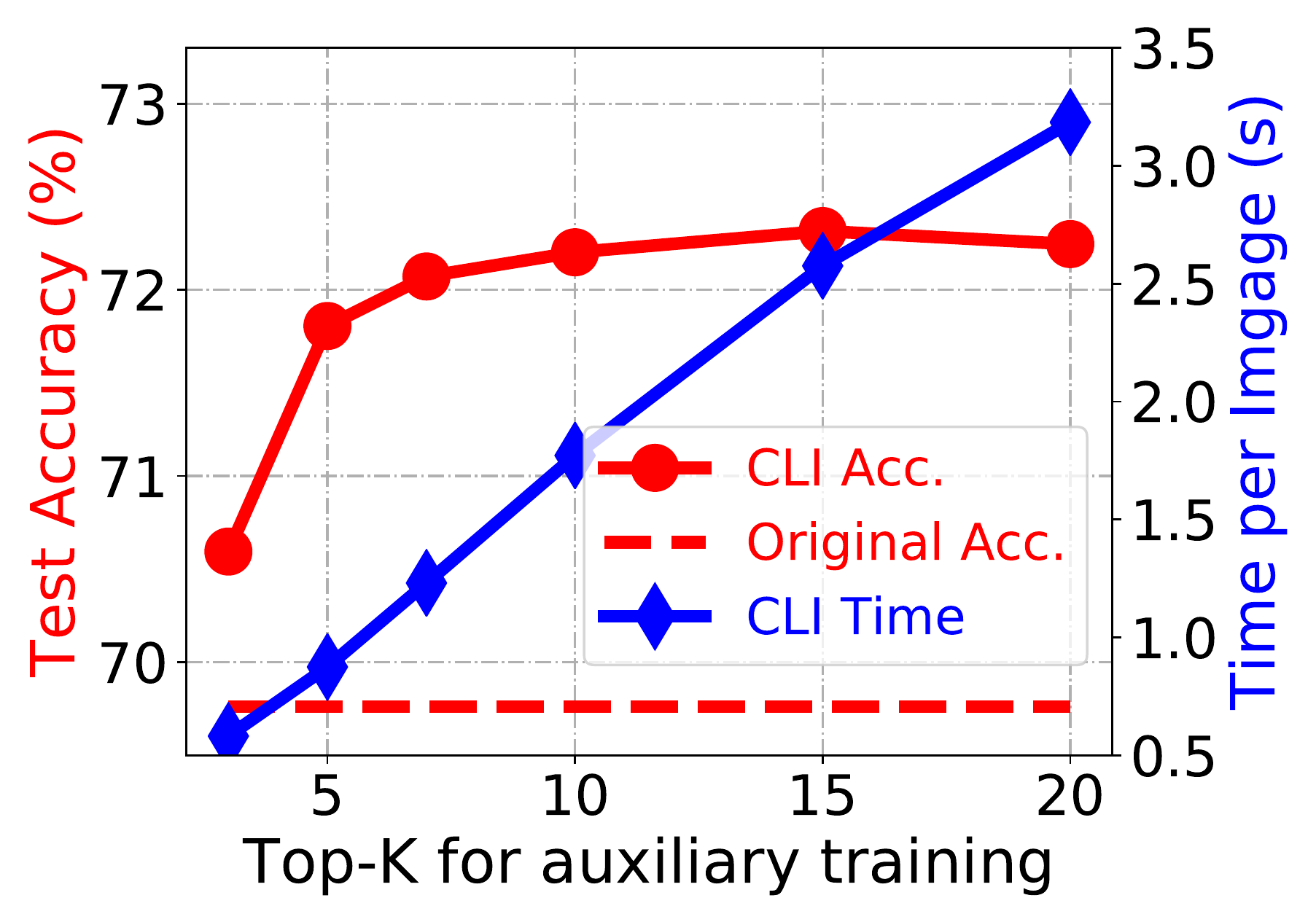}
    \caption{Effects of top-$K$ for auxiliary training.}
    \label{fig:ablation_top_k}
  \end{subfigure}
  \hspace{-0.1in}
  \begin{subfigure}{0.25\linewidth}
    \includegraphics[width=1.0\linewidth]{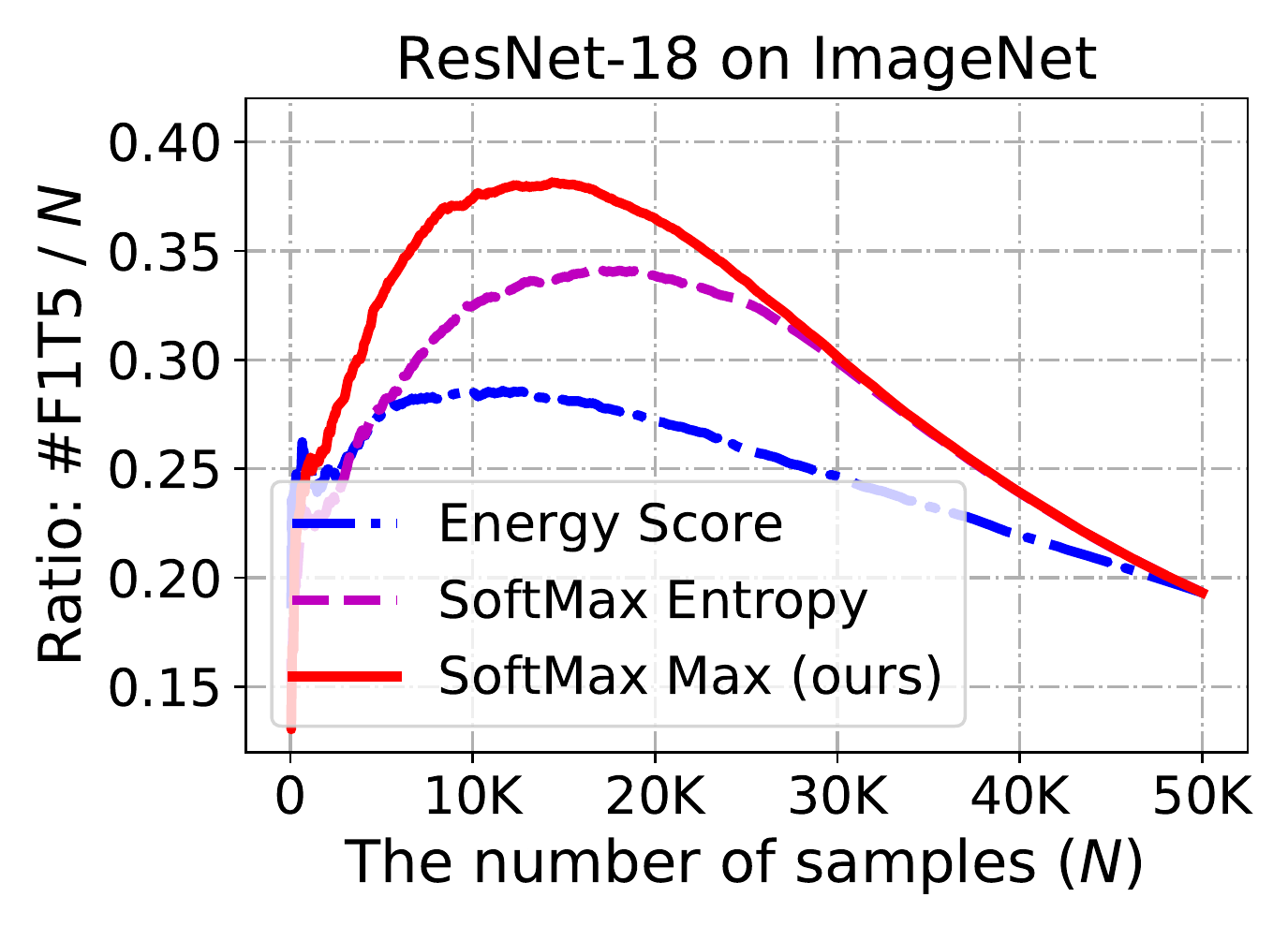}
    \caption{Conditions for sample detection.}
    \label{fig:ablation_diff_condition}
  \end{subfigure}
  \vspace{-0.1in}
  \caption{Ablations on ImageNet pre-trained ResNet-18. In Figure (d), \textbf{Ratio} is calculated as (the number of false top-1 but true top-5 samples (\#F1T5))/(the total number of samples, \ie~the x-axis), and samples in x-axis are sorted via confidence scores.}
  \label{fig:ablations_sum}
\end{figure*}

\paragraph{Number of clusters $Q$.} We evaluate \CLI with different  $Q$, selected from \{200, 300, 400, 500, 600, 700, 800\}, based on ResNet-18. From Figure~\ref{fig:ablations_sum} (a), even with the minimum $Q$ of 200, \CLI improves the top-1 accuracy from 69.76\% to 71.64\% (+1.88\%). When the cluster number equals 500, \CLI achieves the best accuracy of \textbf{73.04\% (+3.28\%)}. In general, with the increase of the cluster number, the performance gain increases, \ie~the cluster number ($\geq500$) achieves better improvement than that of $<500$. These results indicate that a large cluster number benefits the performance of CLI, but the average inference time (called TPI) will increase accordingly.

\paragraph{Threshold $\epsilon$ in Eqn.~(\ref{eq:ood_detector}).}
We evaluate \CLI with different $\epsilon$, selected from \{0.2, 0.3, 0.4, 0.5, 0.6, 0.7, 0.8, 0.9, 1.0\}, based on ResNet-18. From Figure~\ref{fig:ablations_sum} (b), the performance of \CLI improves with the increase of $\epsilon$ (for $\epsilon \leq0.8$). The reason is that a large $\epsilon$ leads to more test images being processed by CLI, and more images will be processed from erroneous to correct. When $\epsilon >0.8$, the performance of \CLI tends to converge, this is because most of the images' original predictions have already been corrected, and thus the improvement becomes stable. It is worth noting that the average inference time (called TPI) for different $\epsilon$ are similar, since the cluster number $Q$ of them is the same. However, with a small $\epsilon$, more images will be predicted without auxiliary training, which is more efficient.

\paragraph{Different top-$K$ for auxiliary training.} 
We evaluate our \CLI with different $K$ values,  selected from \{3, 5, 7, 10, 15, 20\}, for auxiliary training set construction in Eqn.~(\ref{eq:d_aux}). As shown in Figure~\ref{fig:ablations_sum} (c), the performance of \CLI improves with the increase of $K$, but the improvement goes to stable after $K\geq 7$. Moreover, it is worth noting that when $K<5$, the performance improvement achieved by \CLI is quite smaller than that of $K\geq 5$. One possible reason is originated from the K-means clustering step of \CLI, where we use the cluster center to build the auxiliary training set $\mD_{aux}$. Specifically, when $K$ is small (\eg~$K=3$) and there are many samples in one cluster, it is difficult for $\mD_{aux}$ to cover the ground truth class of all test samples within a cluster. In this sense, the model updated on $\mD_{aux}$ will lose the ability to correct samples with wrong top-1 predictions.

\paragraph{Condition for low-confident sample detection.}
We compare our softmax-max score with another two conditions (\ie~Energy score~\cite{liu2020energy} and softmax entropy) for low-confident sample detection.
As described in Section~\ref{sec:selection}, we seek to find more samples that have wrong top-1 but correct top-$K$ predictions, so that \CLI can correctly adjust the top-1 via top-$K$. Therefore, we compute each above score for all test samples and then sort all samples according to the computed score. Then, we define a specific metric for evaluation, \ie~the number of top-1 false but top-5 true images divided by the number of all images (namely, F1T5 Ratio). 
From Figure~\ref{fig:ablations_sum} (d), our softmax-max score achieves a higher ratio than another two conditions, indicating that this condition chooses more top-1 false but top-5 true images.

\begin{table}[t]
\newcommand{\tabincell}[2]{\begin{tabular}{@{}#1@{}}#2\end{tabular}}
 \begin{center}
 \begin{threeparttable}
    \resizebox{1.0\linewidth}{!}{
 	\begin{tabular}{cccc|cc}\toprule
 	Layer1 & Layer2 & Layer3 & Layer4 & ResNet-18  & ResNet-50 \\
 	\midrule
          &  &  & $\checkmark$ & 71.78\% (+2.02\%) & 77.56\% (+1.43\%)\\
          &  & $\checkmark$ & $\checkmark$ & 72.11\% (+2.35\%) & 77.68\% (+1.55\%) \\
          & $\checkmark$ & $\checkmark$ & $\checkmark$ & \textbf{72.28\% (+2.52\%)} & \textbf{77.72\% (+1.59\%)} \\
         $\checkmark$ & $\checkmark$ & $\checkmark$ & $\checkmark$ & 72.20\% (+2.44\%) & 77.65\% (+1.52\%)\\
    \bottomrule
	\end{tabular}
	}
	 \end{threeparttable}
	 \end{center}
	 \vspace{-0.15in}
 \caption{Ablation on the number of layers used for auxiliary training in \CLI. We report the test accuracy on ImageNet.
}
	 \label{tab:ablation_num_layers}
\end{table}

\paragraph{Number of auxiliary training layers.} 
In our main experiments, we perform auxiliary training on all layers of the model. However, as described in Section~\ref{sec:training_details},  training only partial layers of the model is also feasible. This helps to improve the efficiency of auxiliary training and reduce GPU memory consumption. Here, we evaluate the effects of the number of auxiliary training layers in \CLI. We conduct experiments on ResNet-18\&50. As ResNets have 4 layer groups (namely, Layer-1, Layer-2, Layer-3, and Layer-4), we report the results of different layer numbers in Table~\ref{tab:ablation_num_layers}. From the results, training Layer-2, Layer-3 and Layer-4 achieves the best performance. 
Compared with it, training the whole model or training Layer-3 and Layer-4 achieves comparable performance. These results verify that training partial layers of the model in auxiliary training is enough to correct the model's wrong top-1 predictions.

\section{Conclusions}
In this paper, we have proposed a Closed-Loop Inference (\CLI) method that can be considered as a general approach to boost the classification ability of any pre-trained model. To this end, we devise a sample detection condition to filter images whose predictions are not confident enough. For the filtered images, we construct an auxiliary training task based on the original top-$K$ predicted classes, and then make the final prediction using the newly auxiliary trained model. Moreover, we exploit the K-means clustering technique to achieve the goal of auxiliary training once and inference many, thereby improving the overall efficiency of \CLI. Experimental results show that our \CLI boosts the predictive performance on both in-distribution and out-of-distribution test samples. In the future, it would be interesting to extend our method to a broader of prediction tasks, such as object detection and semantic segmentation.

~\\
\noindent\textbf{Potential limitations.} 
Similar to previous test-time learning methods, \CLI sacrifices the inference efficiency for more accurate predictions, \eg~2.71\% accuracy gain on ImageNet at the cost of 1.69 seconds per image ($\epsilon\small{=}0.8$ in Figure~\ref{fig:ablations_sum} (b)). This is highly favorable for application scenarios where the inference accuracy is more critical than the computational efficiency, such as medical image analysis~\cite{anthimopoulos2016lung,minaee2020deep}. For those latency-sensitive applications, we would like to point out that the efficiency of CLI could be improved from two aspects. Firstly, certain hyper-parameters can be tweaked to adjust the accuracy-efficiency trade-off, as illustrated in Section~\ref{sec:ablations_all} and Supplementary. Secondly, auxiliary learning tasks can be constructed on a compact subset of training samples, following techniques including Prototype \cite{snell2017prototypical} and Grad-Match \cite{killamsetty2021grad}. Nevertheless, it still remains as an open question on how to further improve the training efficiency, which we leave to our future work.

{
\small  
\bibliographystyle{ieee_fullname}
\bibliography{egbib}
}

\clearpage

\renewcommand\thesection{\Alph{section}}
\renewcommand\thefigure{\Alph{figure}}
\renewcommand\thetable{\Alph{table}}
\setcounter{section}{0}
\setcounter{figure}{0}
\setcounter{table}{0}

\onecolumn{
\title{
\begin{center}
\textbf{
\Large{Supplementary Materials for ``Boost Test-Time Performance \\ with Closed-Loop Inference"}
}
\end{center}
}
}

\vspace{25pt}

We organize our supplementary as follows. In Section~\ref{sec:visiualization_images}, we provide extra experimental results on ImageNet~\cite{deng2009imagenet} (in-distribution test samples). In Section~\ref{sec:details_ttt_cli}, we provide additional experimental details on ImageNet-C~\cite{hendrycks2019benchmarking} (out-of-distribution test samples). In Section~\ref{sec:ablations_supp}, we conduct ablation studies to verify the effects of hyper-parameters and components in our \CLI. 

~\\
\noindent\textbf{Evaluation metrics.} 1) \textbf{\#F2T}: the number of test images whose top-1 predictions are corrected from \textbf{f}alse \textbf{t}o \textbf{t}rue (F2T). 2) \textbf{\#T2F}: the number of test images whose top-1 predictions are corrected from \textbf{t}rue \textbf{t}o \textbf{f}alse (T2F). The percentage (\#\%) for F2T/T2F is calculated by the number of F2T/T2F samples divided by the total amount of test samples (50,000 for ImageNet). 
3) \textbf{F2F} and \textbf{T2T} have the similar meanings. 
4) \textbf{Time per Image (TPI)}: the average run-time for processing one image via \CLI, calculated by the total run time (on a single Tesla-V100) of our algorithm dividing the number of test samples (50,000).

\section{More Results on In-Distribution Test Samples}\label{sec:visiualization_images}
\paragraph{Visualization of low-confident images processed by \CLI.}
We visualize four categories of ImageNet~\cite{deng2009imagenet} test images processed by our \CLI in Figure~\ref{fig:Visualization_supp}. 
1) \textbf{F2T} images mainly belong to fine-grained classification, such as (King crab v.s. Dungeness crab) and (Wok v.s. Frypan). 
2) For \textbf{F2F} images, they also belong to the fine-grained classification (Left 1\&2), or images themselves are hard to be classified (Right 1). Although the \CLI prediction is still false, it is closer to the ground truth. For example, the Left 1 image's original and \CLI predictions are Italian greyhound and Garfish, where the latter one is closer to the ground truth of Hammerhead shark.
3) \textbf{T2F} images also contain two types. The first is images with multiple objects (Left 1\&2) but the ground truth only annotates one of them. Here, \CLI adjusts the original prediction (label) to be another, which in fact also belongs to this image. The second type is also the fine-grained classification (Right 1\&2).
4) For \textbf{T2T} images, the predictions with \CLI become more confident, \eg~a Joystick image (Right 2) with a confidence score from 0.329 to 0.928.

\definecolor{textblue}{RGB}{47,137,203}
\begin{figure*}[h!]
\centering
\includegraphics[width=1.\linewidth]{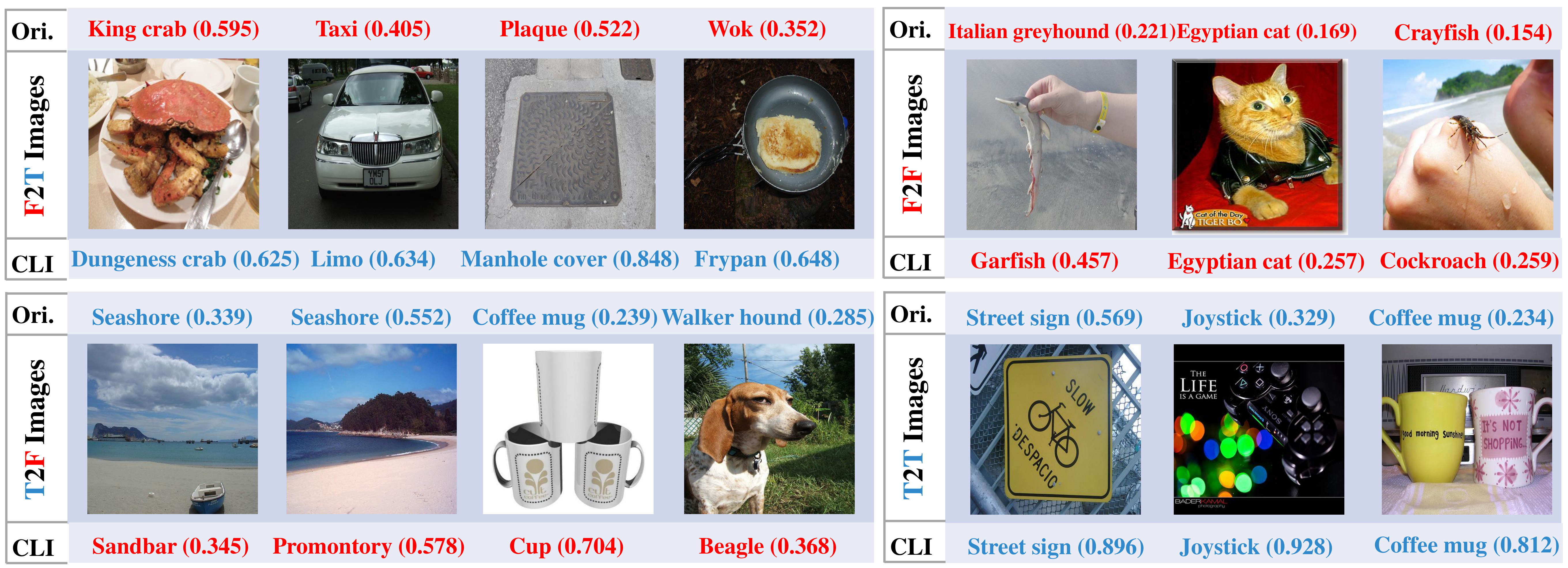}
\vspace{-0.18in}
\caption{Low-confident images processed by \CLI with ResNet-18. We show four types (F2T, F2F, T2F, and T2T) images' original (Ori.) and final \CLI predicted class with the associated confidence score. The \textcolor{red}{red} and \textcolor{textblue}{blue} colors denote the predicted class is \textcolor{red}{false} and \textcolor{textblue}{true}.}
\label{fig:Visualization_supp}
\end{figure*}

\paragraph{\CLI on various networks.} In this section, we provide a detailed version of Table 1 in the main paper. Here, we further report the number of \textbf{F2T} and \textbf{T2F} images processed by \CLI on different pre-trained models. From the results, at the same time as our \CLI correcting the prediction of test images from false to true, it also adjusts a smaller number of predictions from true to false. One possible reason is that the model forgets some previous knowledge during the auxiliary training phase. How to reduce \#T2F is still an open question, which we leave to our future work. 

\begin{table*}[t!]
\newcommand{\tabincell}[2]{\begin{tabular}{@{}#1@{}}#2\end{tabular}}
 \begin{center}
 \begin{threeparttable}
    \resizebox{0.88\linewidth}{!}{
 	\begin{tabular}{l|cc|c|cc|c}\toprule
 	Model  & \#Params. & \#FLOPs & Original Acc. & \#F2T ($\uparrow$) & \#T2F ($\downarrow$) & \CLI Acc.  (ours) \\
        \midrule
         ResNet-18~\cite{he2016deep} & 11.7M & 1.82B & 69.76\% & 2,491 (4.98\%) & 1,269 (2.54\%)& \textbf{72.20\% (+2.44\%)}\\
         ResNet-50~\cite{he2016deep} & 25.6M & 4.11B & 76.13\% & 1,925 (3.85\%) & 1,165 (2.33\%) & \textbf{77.65\% (+1.52\%)}\\
         ResNet-101~\cite{he2016deep} & 44.6M & 7.83B & 77.37\% & 1,778 (3.56\%) & 1,169 (2.34\%)& \textbf{78.59\% (+1.22\%)}\\
         ResNet-152~\cite{he2016deep} & 60.2M & 11.56B & 78.31\% & 1,572 (3.14\%) & 1,022 (2.04\%)& \textbf{79.41\% (+1.10\%)}\\
         ResNeXt-101~\cite{xie2017aggregated} & 88.8M & 16.48B & 79.31\% & 1,206 (2.41\%) & 917 (1.83\%) & \textbf{79.89\% (+0.58\%)}\\
         ResNeXt-101-WSL~\cite{mahajan2018exploring} & 88.8M & 16.48B & 82.69\% & 1,073 (2.15\%)& 784 (1.57\%)& \textbf{83.27\% (+0.58\%)}\\
         \midrule
         DenseNet-121~\cite{huang2017densely} & 8.0M & 2.87B & 74.43\% & 2,235 (4.47\%) & 1,253 (2.51\%)& \textbf{76.40\% (+1.96\%)}\\
         DenseNet-169~\cite{huang2017densely} & 14.2M & 3.40B & 75.60\% & 2,077 (4.15\%)& 1,125 (2.25\%)& \textbf{77.50\% (+1.90\%)}\\
         Inception-V3~\cite{szegedy2016rethinking} & 23.8M & 5.73B & 77.47\% & 1,876 (3.75\%)& 1,193 (2.39\%)& \textbf{78.84\% (+1.37\%)}\\ 
         \midrule
         ShuffleNet-V2~\cite{ma2018shufflenet} & 2.3M & 0.15B & 69.36\% & 1,776 (3.55\%) & 1,271 (2.54\%)& \textbf{70.37\% (+1.01\%)}\\ 
        MobileNet-V2~\cite{sandler2018mobilenetv2} & 3.5M & 0.31B & 71.88\% & 2,292 (4.58\%)& 1,367 (2.73\%)& \textbf{73.73\% (+1.85\%)}\\ 
         MobileNet-V3~\cite{howard2019searching} & 5.5M & 0.23B & 74.04\% & 1,684 (3.37\%)& 1,102 (2.20\%)& \textbf{75.21\% (+1.17\%)}\\ 
         EfficientNet-B0~\cite{tan2019efficientnet} & 5.3M & 0.39B & 76.13\% & 1,980 (3.96\%)& 1,241 (2.48\%)& \textbf{77.91\% (+1.48\%)}\\
         EfficientNet-B1~\cite{tan2019efficientnet} & 7.8M & 0.70B & 78.38\% & 1,646 (3.29\%)& 1,197 (2.39\%)& \textbf{79.28\% (+0.90\%)}\\
        \bottomrule 
	\end{tabular}
	}
	 \end{threeparttable}
	 \end{center}
	 \vspace{-0.15in}
    \caption{Test accuracy on several classic and state-of-the-art models on ImageNet. A detailed version of Table 1 in the main paper.
    }
    \vspace{-0.15in}
	 \label{tab:main_results_supp}
\end{table*}

\section{More Experimental Details on Out-of-Distribution Test Samples}\label{sec:details_ttt_cli}

\begin{figure*}[h!]
\centering
\includegraphics[width=0.72\linewidth]{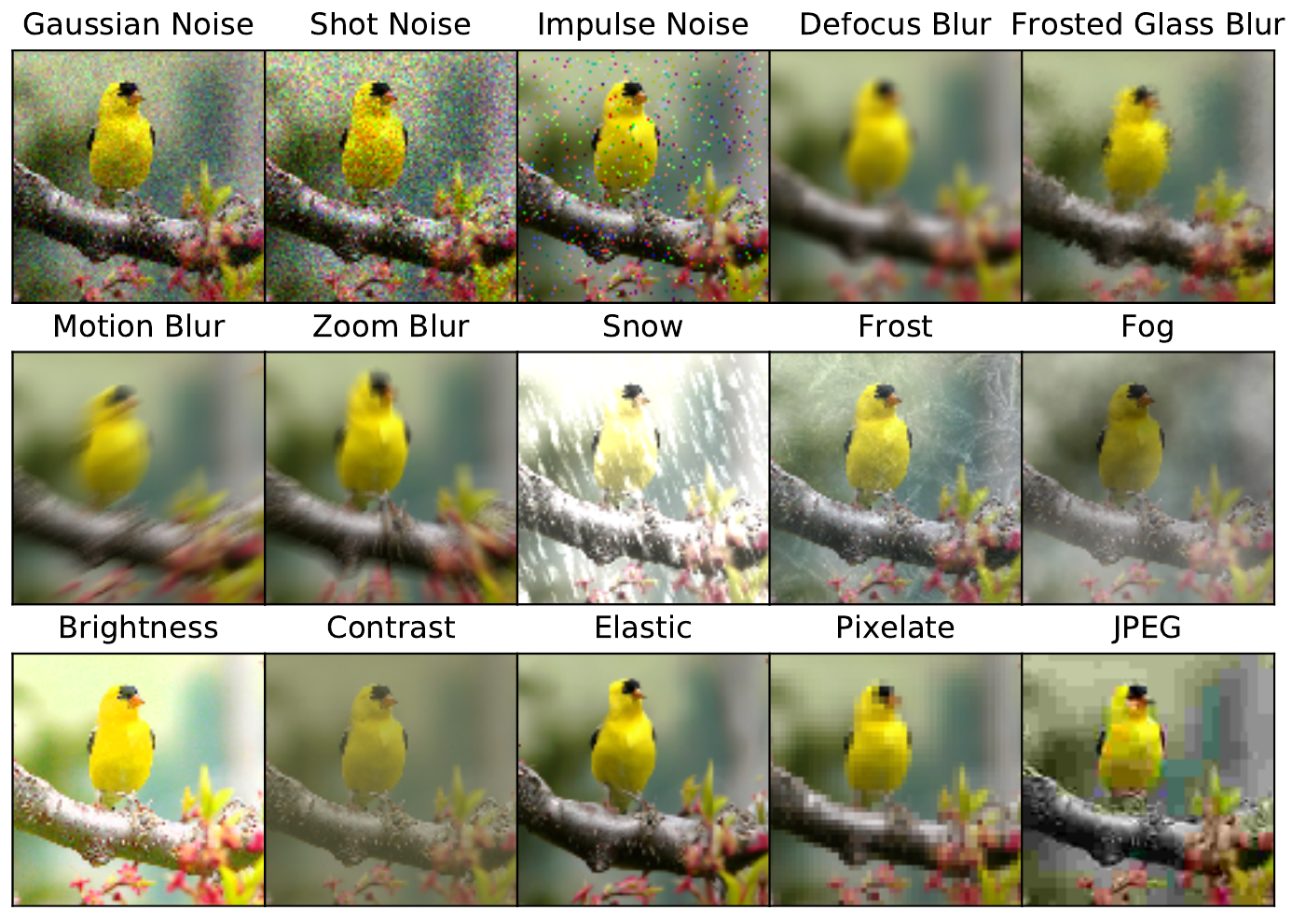}
\vspace{-0.12in}
\caption{Different corruption types in ImageNet-C~\cite{hendrycks2019benchmarking}, taken from the original paper by ImageNet-C~\cite{hendrycks2019benchmarking}.}
\label{fig:corruption_types}
\end{figure*}

\paragraph{Dataset.} In the main paper, we evaluate TTT~\cite{sun2020test}+CLI on ImageNet-C~\cite{hendrycks2019benchmarking} to verify the effectiveness of our \CLI in improving the model performance on out-of-distribution test samples. Here, ImageNet-C is constructed by corrupting the ImageNet~\cite{deng2009imagenet} test set. The corruption (as shown in Figure~\ref{fig:corruption_types}) consists of 15 different types, \ie~Gaussian noise,
shot noise, impulse noise, defocus blur, glass blue, motion blur, zoom blur, snow, frost, fog, brightness, contrast, elastic transformation,
pixelation, and JPEG compression.

\paragraph{Implementation details of TTT+\CLI.}
In the main paper, we apply our \CLI to a previous state-of-the-art method, called test-time training (TTT)~\cite{sun2020test}, to further improve its performance on out-of-distribution test samples. In the original paper of TTT, it requires the model learned via both the rotation prediction~\cite{gidaris2018unsupervised} and classification loss during the training phase. However, we aim to boost the performance of any pre-trained model and do not require any operation during the training phase. For fair comparisons, we modify TTT to make it can also be directly applied to a pre-trained model. Specifically, given a pre-trained model (ResNet-50), we add a new branch (random initialized) from the end of 3rd group of ResNet for the rotation prediction task. We first freeze all original parameters of the pre-trained model and train the newly added branch for 5 epochs on the whole ImageNet training set. Here, we apply an SGD optimizer with an initial learning rate 0.01 and decrease it by cosine annealing~\cite{SGDR}. Then, we take the newly obtained model (with two branches) as the base model to perform test-time training on all test data for one gradient step. Following TTT, we use an SGD optimizer with a learning rate of 0.001. Last, we apply \CLI to the final model (after performing TTT) to further boost its performance. Moreover, we replace the Group Normalization in TTT with Batch Normalization, which is much more widely used. Interestingly, as shown in Table~\ref{tab:comparison_ttt_our_impl}, TTT with our modifications often achieves better performance than its original implementation.

\begin{table*}[t!]
\newcommand{\tabincell}[2]{\begin{tabular}{@{}#1@{}}#2\end{tabular}}
 \begin{center}
 \begin{threeparttable}
    \resizebox{1.0\linewidth}{!}{
 	\begin{tabular}{c|l|ccc|cccc|cccc|cccc}
 	\multicolumn{2}{c}{} & \multicolumn{3}{c}{Noise} & \multicolumn{4}{c}{Blur} & \multicolumn{4}{c}{Weather} & \multicolumn{4}{c}{Digital} \\
 	\toprule
 	Severity & Method & Gauss. & Shot & Impulse & Defocus & Glass & Motion & Zoom & Snow & Frost & Fog & Britght & Contrast & Elastic & Pixel & JPEG \\
 	\midrule
 	\multirow{2}{*}{\tabincell{c}{Level-1}}
        & TTT~\cite{sun2020test}  & 62.51 & 62.60 & 60.57 & 58.22 & 60.29 & 62.81 & 56.06 & 58.26 & 59.51 & 63.70 & 69.01 & 67.06 & 65.00 & 66.24 & 63.79 \\
        & TTT~\cite{sun2020test}$^\dagger$             & 66.81&	66.83&	63.43&	61.77&	63.71 &	63.23&	58.42&	57.43&	62.87 &	68.28 &	73.21&	69.98 &	69.31 &	69.70 &	66.54 \\
    \midrule
 	\multirow{2}{*}{\tabincell{c}{Level-2}}
        & TTT~\cite{sun2020test}  & 58.01 & 57.50 & 55.66 & 52.29 & 46.71 & 57.30 & 51.81 & 49.61 & 43.39 & 61.76 & 68.17 & 65.73 & 50.58 & 65.32 & 61.68 \\
        & TTT~\cite{sun2020test}$^\dagger$ & 62.26&	61.60&	58.37&	55.30&	47.76&	54.23&	55.63 &	48.83 &	48.53 &	66.10 &	72.33&	68.32 &	55.56 &	68.69 &	63.46 \\
    \midrule
 	\multirow{2}{*}{\tabincell{c}{Level-3}}
         & TTT~\cite{sun2020test}  & 51.53 & 51.15 & 51.67 & 42.30 & 32.86 & 49.31 & 49.28 & 48.78 & 29.74 & 59.02 & 66.73 & 62.41 & 63.75 & 61.64 & 60.08 \\
         & TTT~\cite{sun2020test}$^\dagger$ & 55.00 &	55.19 &	54.30 &	46.11 &	32.54 &	36.79&	52.60&	49.26 &	34.97&	63.63 &	70.78 &	64.93 &	66.69 &	63.98 &	61.43 \\
        \bottomrule
	\end{tabular}
	}
	 \end{threeparttable}
	 \end{center}
	 \vspace{-0.1in}
    \caption{Comparison between original results of TTT~\cite{sun2020test} and our implementation. We report the test accuracy (\%) on ImageNet-C with different severity-levels. The backbone model is ResNet-50. $^\dagger$ denotes our implementation.
    }
    \label{tab:comparison_ttt_our_impl}
\end{table*}

\section{More Ablations}\label{sec:ablations_supp}
In this section, we conduct additional ablation studies to verify the effects of hyper-parameters and components in \CLI. The overall hyper-parameter settings and auxiliary training details are the same as the main paper. To be specific, we set the cluster number $Q$ to 400, the number of auxiliary training epochs to 5, the proportion of auxiliary training set to 1.0, and threshold $\epsilon$ for  low-confident sample selection to 0.7. We (auxiliary) train the whole network with the supervised contrastive loss. For each experiment, we only adjust one hyper-parameter and keep others fixed. All experiments are conducted on ImageNet pre-trained ResNet-18.

\begin{figure}[h!]
\centering
\includegraphics[width=0.4\linewidth]{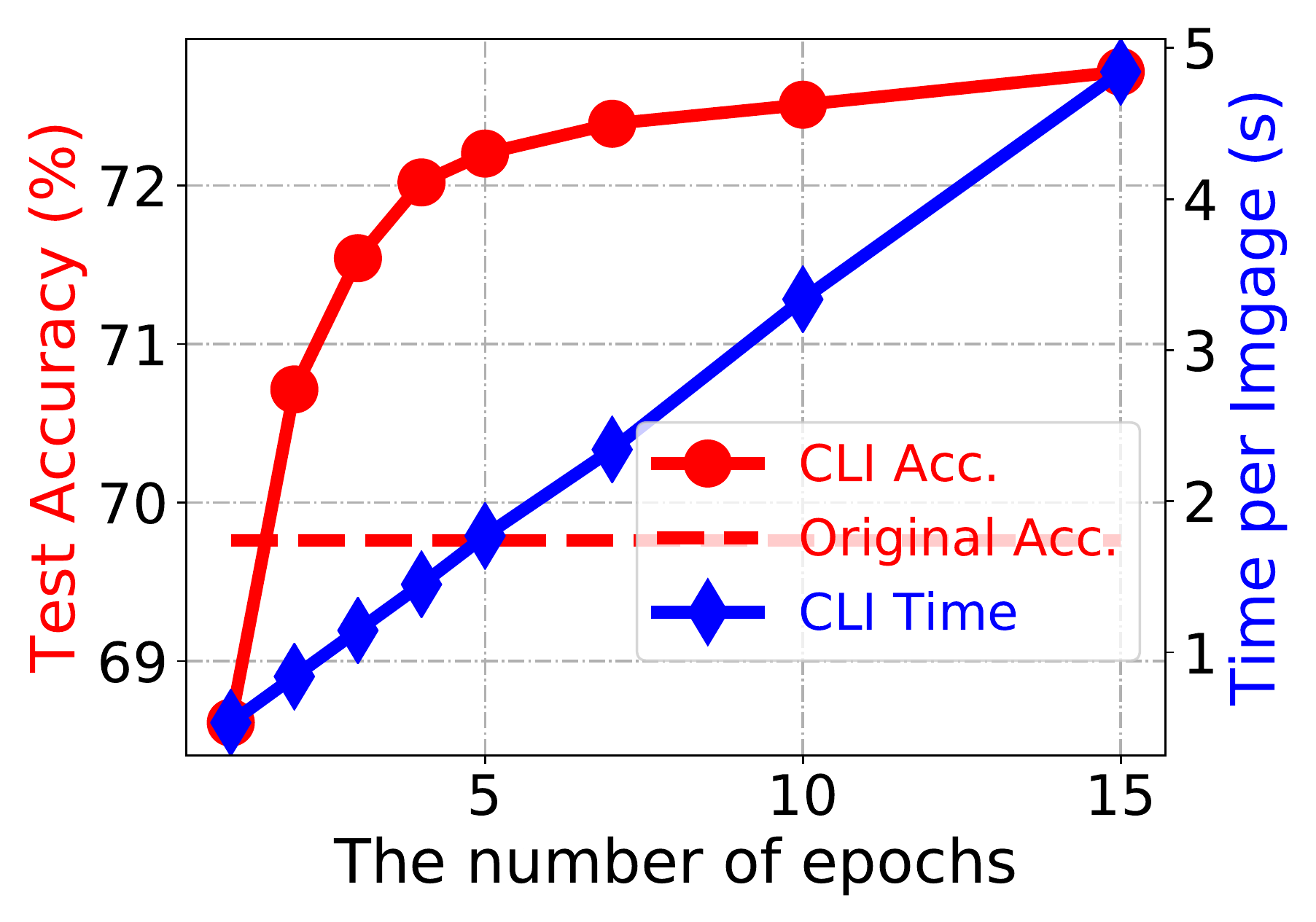}
\vspace{-0.1in}
\caption{Effects of auxiliary training epochs. Experiments are conducted on ImageNet pre-trained ResNet-18.}
\label{fig:ablations_epochs}
\end{figure}

\paragraph{Number of auxiliary training epochs.} 
We evaluate our \CLI with different numbers of auxiliary training epochs from \{1, 2, 3, 4, 5, 7, 10, 15\} on ResNet-18. From Figure~\ref{fig:ablations_epochs}, the performance of \CLI improves with the increase of training epochs, but the improvement speed slows down when epochs$>$5. As more epochs lead to a longer inference time per image (called TPI), one can choose a suitable number of epochs for the trade-off between performance and efficiency. In our main experiments, we set the training epochs to 5 and achieve 2.44\% top-1 accuracy improvement with 1.77 seconds of TPI.

\begin{figure}[h!]
\centering
\includegraphics[width=0.4\linewidth]{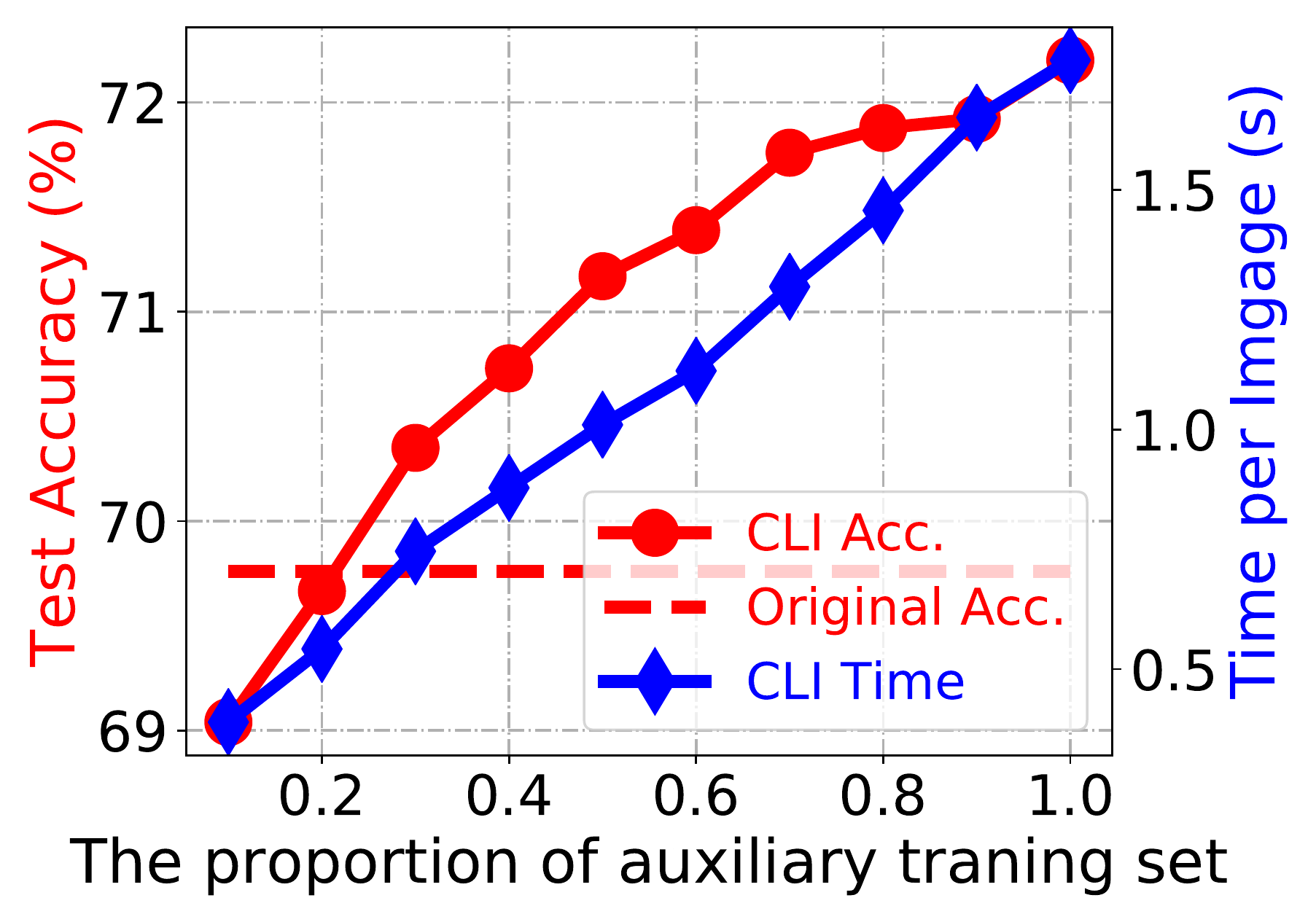}
\vspace{-0.1in}
\caption{Effects of the proportion of auxiliary training set. Experiments are conducted on ImageNet with ResNet-18.}
\label{fig:ablations_subset_ratio}
\end{figure}

\paragraph{Proportion of auxiliary training set.} 
We evaluate \CLI with different proportion sizes of auxiliary training set from \{0.1, 0.2, 0.3, 0.4, 0.5, 0.6, 0.7, 0.8, 0.9, 1.0\} on ResNet-18. Results in Figure~\ref{fig:ablations_subset_ratio} show that more auxiliary training data help the \CLI to better adjust the model's original prediction from erroneous to correct. This demonstrates the importance of data in \CLI. Nevertheless, one can choose the most informative data for auxiliary training, and thus improve the algorithm efficiency by reducing the auxiliary training costs. We leave this to our future work.

\paragraph{Comparison with \CLI-Online.}
In this section, we compare our \CLI with its online version, which processes test images one by one and does not need the clustering step of Eqn.~(5) in the main paper. Due to the high run-time cost of \CLI-Online (about 1.75 minutes per image), we only compare them with threshold $\epsilon\leq0.6$. As shown in Table~\ref{tab:ablation_online}, \CLI achieves comparable performance with \CLI-Online but with much lower TPI, i.e., 1.71 seconds per image when $\epsilon=0.6$. These results verify the necessity of clustering on improving the algorithm efficiency.

\begin{table}[h!]
\newcommand{\tabincell}[2]{\begin{tabular}{@{}#1@{}}#2\end{tabular}}
 \begin{center}
 \begin{threeparttable}
    \resizebox{0.55\linewidth}{!}{
 	\begin{tabular}{c|c|c|c}\toprule
 	 Model & $\epsilon$ in Eqn.~(3) & \CLI-Online Acc. & \CLI Acc. \\
 	\midrule
         \multirow{4}{*}{\tabincell{c}{ResNet-18 \\ (69.76\%)}} & 0.30 & 70.47\% (+0.71\%)  & 70.29\% (+0.53\%)  \\
         & 0.40 & 70.96\% (+1.20\%)  & 70.74\% (+0.98\%)  \\
         & 0.50 & 71.45\% (+1.69\%)  & 71.37\% (+1.61\%)  \\
         & 0.60 & 71.86\% (+2.10\%) & 71.80\% (+2.04\%)  \\
    \bottomrule
	\end{tabular}
	}
	 \end{threeparttable}
	 \end{center}
	 \vspace{-0.1in}
\caption{Comparison with \CLI-Online regarding different thresholds $\epsilon$ (in Eqn.~(3) of the main paper) on ImageNet. }
\label{tab:ablation_online}
\end{table}

\end{document}